\documentclass[10pt]{article} %
\usepackage[nonatbib]{neurips_widetext}
\usepackage[utf8]{inputenc} %
\usepackage[T1]{fontenc}    %
\usepackage{url}            %
\usepackage{booktabs}       %
\usepackage{amsfonts}       %
\usepackage{nicefrac}       %
\usepackage{microtype}      %
\usepackage{lipsum}         %
\usepackage{graphicx}
\usepackage{doi}
\usepackage{amsmath}
\usepackage{amssymb}
\usepackage{amsthm}
\usepackage{bbding}
\usepackage{multirow}
\usepackage{subcaption}
\usepackage{algorithm}
\usepackage{algorithmic}
\usepackage{cleveref}       %
\usepackage{enumitem}
\usepackage{makecell}   %
\usepackage{color}
\usepackage{hyperref}       %
\usepackage{minitoc}
\usepackage[numbers]{natbib}   %
\title{Accurate Neural Network Pruning Requires \\ Rethinking Sparse Optimization}

\author{
    Denis Kuznedelev\thanks{These authors contributed equally. Author order was determined by experimental load (highest first).} \\
    Skoltech \& Yandex\\
    \texttt{denis.kuznedelev@skoltech.ru} \\
    \And
    Eldar Kurtic\footnotemark[1] \\
    IST Austria \\
    \texttt{eldar.kurtic@ista.ac.at} \\
    \And
    Eugenia Iofinova\footnotemark[1] \\
    IST Austria \\
    \texttt{eugenia.iofinova@ista.ac.at} \\
    \And
    Elias Frantar \\
    IST Austria \\
    \texttt{elias.frantar@ista.ac.at} \\
    \And
    Alexandra Peste \\
    IST Austria \\
    \texttt{alexandra.peste@ista.ac.at} \\
    \And
    Dan Alistarh \\
    IST Austria \& Neural Magic \\
    \texttt{dan.alistarh@ista.ac.at} \\
}

\begin{document}
\doparttoc %
\faketableofcontents %

\maketitle

\maketitle

\begin{abstract}
Obtaining versions of deep neural networks that are both highly-accurate and highly-sparse %
is one of the main challenges in the area of model compression, and several high-performance pruning techniques have been investigated by the community. 
Yet, much less is known about the interaction between sparsity and the standard stochastic optimization techniques used for training sparse networks, and most existing work uses standard dense schedules and hyperparameters for training sparse networks.  
In this work, we examine the impact of high sparsity on model training using the standard computer vision and natural language processing sparsity benchmarks. 
We begin by showing that using standard dense training recipes for sparse training is suboptimal, and results in under-training. We provide new approaches for mitigating this issue for both sparse pre-training of vision models (e.g. ResNet50/ImageNet) and sparse fine-tuning of language models (e.g. BERT/GLUE), achieving state-of-the-art results in both settings in the high-sparsity regime, and providing detailed analyses for the difficulty of sparse training in both scenarios. 
Our work sets a new threshold in terms of the accuracies that can be achieved under high sparsity, and should  inspire further research into improving sparse model training, to reach higher accuracies under high sparsity, but also to do so efficiently.
\end{abstract}

\section{Introduction}

The difficulty of finding deep neural networks (DNNs) that are both \emph{accurate and sparse}, 
i.e., closely match the accuracy of dense models while having a large majority of their weights set to zero, 
is one of the main challenges in the area of model compression. 
On the conceptual side, this challenge connects to fundamental questions related to the 
\emph{Lottery Ticket Hypothesis (LTH)}~\cite{frankle2018lottery, frankle2019stabilizing}, which posited that such sparse masks exist, 
and that, in some cases, they can even allow accurate training of sparse models \emph{from scratch}, that is, 
applying the sparsity mask at initialization.
On the practical side, obtaining highly-sparse and accurate networks can lead to significant practical speedups, both for inference~\cite{deepsparse} and training~\cite{nikdan2023sparseprop}. 

In this work, we focus on the challenge of obtaining accurate DNNs in the high-sparsity regime, and investigate the barriers to obtaining \textbf{highly-sparse} and \textbf{highly-accurate} variants of DNNs  for standard vision and language tasks.
We mainly focus on two tasks that are, arguably, the standard benchmarks for sparsity in vision and language, respectively: 
image classification using the ResNet50 model~\cite{he2016deep} on the ImageNet-1K dataset~\cite{imagenet}, e.g.~\cite{hoefler2021sparsity, 2017-dong, gale2019state, evci2020rigging, singh2020woodfisher, savarese2021winning, peste2021ac}, 
and language modelling using the BERT-base model~\cite{devlin2018bert} on the GLUE benchmark datasets~\cite{wang2018glue}, e.g.~\cite{2020-sanh, hoefler2021sparsity, kurtic2022gmp, kurtic2022optimal}. 
Roughly, for both benchmarks, it is known that sparsities lower than 90\% can be achieved with approximately 1\% accuracy loss relative to the original dense model, 
but accuracy rapidly decreases in the 90-95\% range~\cite{hoefler2021sparsity, evci2020rigging}, and that decreases are drastic at higher ($\geq 95\%$)  sparsities~\cite{singh2020woodfisher, kurtic2022optimal}. 
In this paper, we investigate the reasons behind this accuracy loss due to sparsity, mainly targeting \emph{high sparsity}, i.e. sparsities between 90\% and 99\%, studying the difficulty of obtaining accurate models in this range. 

\paragraph{Contribution.}  
We begin from the observation that, when training sparse models from scratch, following standard \emph{dense training} schedules, \emph{sparse models show clear evidence of undertraining}: both their accuracy and loss fail to saturate, and their output continues to have high entropy. 
This finding suggests that maximization of the accuracy of sparse models requires longer training than the dense optimization recipes adopted in most of the work on model sparsification.

Motivated by this observation, we propose a combination of techniques which can mitigate the inherent difficulty of sparse training. As a consequence, we  significantly improve on the best currently-known sparsity-accuracy trade-offs on standard sparsity benchmarks for both image classification and language modelling. 
More precisely, we obtain, for the first time, highly-accurate sparse versions of ResNet50, such as a 90\%-sparse model with 78.5\% Top-1 accuracy, a 95\%-sparse model with 77.7\% Top-1 accuracy, and a 98\%-sparse model with 75.2\% Top-1 accuracy. In addition, we show that stable results can be obtained even for extreme  sparsities (e.g., 99\%). 
For language models, we show that on the most challenging tasks, as measured by the drop in accuracy relative to the dense model, we can improve results by 3 points in accuracy relative to the current state-of-the-art results at 90\% sparsity. 
We arrive at these results as follows:  

\begin{itemize} 
\item We perform an analysis of the output and training characteristics of models trained using current state-of-the-art techniques, relative to their dense counterparts. First, we show that sparse DNNs obtained via many current techniques behave similarly to \emph{undertrained dense models}: specifically, they tend to have high output entropy (alternatively, low ``output confidence''), which correlates with  their reduced accuracy. 

\item This analysis provides clear evidence that optimizing \emph{sparse models} is more difficult than standard \emph{dense} optimization~\cite{evci2019difficulty}. This observation stands in  contrast to the fact that most current sparsification techniques use standard \emph{dense} training recipes for fine-tuning and recovery. We exploit this insight to obtain state-of-the-art accuracy for sparse models in two popular scenarios: \emph{sparse pretraining}, i.e. training sparse models from scratch, and \emph{sparse transfer}, i.e. optimizing a sparse pretrained model onto a target transfer task. 

\item In the \emph{sparse pretraining} scenario, illustrated by the standard task of obtaining a highly-sparse ResNet50 model on the ImageNet dataset, we show that we can circumvent the difficulty of sparse training by adopting a variant of the Alternating Compressed/Decompressed (AC/DC) algorithm~\cite{peste2021ac} for training sparse DNNs, which has guarantees for sparse recovery. Specifically, we show that, by scaling the algorithm's runtime, we can obtain state-of-the-art results for sparse pretraining on ImageNet for ResNet50 and MobileNet models, and reach extremely high sparsities (e.g. 98\% and 99\%) while still obtaining stable results. Moreover, only sparse models benefit from extended training, whereas dense models start to overfit with longer training.%

\item In the \emph{sparse transfer} scenario, popular in language domain, the difficulty of sparse training can manifest itself through both \emph{undertraining} and \emph{overfitting}, depending on the parametrization of the chosen transfer learning recipe, specifically on the training length. 
We address this via a modified version of the \emph{gradual layer unfreezing} approach~\cite{howard2018universal}, tailored towards a \emph{sparse} transfer learning scenario, which allows us to obtain state-of-the-art results in the case of BERT-base transfer on downstream datasets. 

\end{itemize}

\paragraph{Discussion.} 
Overall, our results suggest that the difficulty of obtaining highly-accurate sparse models is closely linked to the difficulty of accurate sparse optimization using current state-of-the-art techniques. 
Specifically, our work improves the best known results on standard sparsity benchmarks, for both sparse pretraining and sparse finetuning, both in terms of absolute accuracy, and accuracy loss relative to the dense baseline. 
Moreover, we observe the following:

\begin{itemize}
    \item Achieving state-of-the-art sparsity-vs-accuracy trade-offs currently requires using significant additional computational complexity and more epochs %
    for training the sparse models, relative to the best known dense training methods. In turn, this suggests that sparse optimization may be inherently harder than its dense counterpart. 

    \item Reaching high validation accuracy for sparse models is strongly linked to reaching low training loss, which occurs at a slower rate for sparse models in the case of SGD-based optimization. At the same time, we do observe overfitting behavior (decrease of validation accuracy w.r.t. increased training time), especially at lower sparsities. 
    
    \item To further investigate the hardness of sparse optimization, we perform an analysis of the loss landscape of accurate sparse networks both in terms of sharpness and loss interpolation / mode connectivity. We observe that achieving highly-accurate sparse networks from initialization requires overcoming multiple loss barriers, and that sparsity mask exploration may be a key ingredient for overcoming these barriers. 

    \item In addition, we investigate the relationship between standard hyperparameters such as weight decay, on the one hand, and sparsity structure, on the other. We find that careful setting of weight decay is critical for accurate sparsity, and that weight decay additionally induces (partial) structured sparsity in highly-sparse models. This provides a first explanation to the emergence of structured sparsity in unstructured sparse networks, which has been observed previously~\cite{peste2021ac, iofinova2022transfer, yin2023dynamic}. 
\end{itemize}

Our results set new accuracy thresholds for sparse models using relatively simple techniques. They should serve as motivation for the community to devise improved \emph{sparsity-aware} optimization techniques, specifically allowing for faster, more efficient accuracy recovery.

\section{Related Work}
\label{sec:related_work}

The goal of most sparsification methods~\cite{hoefler2021sparsity} is to create a DNN that is as accurate as possible, while maximizing sparsity. 
This goal can be achieved via different strategies: for instance, \emph{post-training sparsification methods} assume a \emph{pretrained dense model}, 
from which weights are removed either in a single step (one-shot) or progressively (gradual pruning). 
By contrast, in \emph{sparse training methods}, parameters are pruned from the model during training from scratch, 
either close to initialization~\cite{evci2020rigging, jayakumar2021top, lee2018snip, vanholder2016efficient, schwarz2021powerpropagation}, or progressively as the model is trained~\cite{han2015learning,gale2019state, savarese2021winning}. 
A subset of sparse training methods are \emph{dynamic}, in the sense that weights may be reintroduced during training~\cite{evci2020rigging, peste2021ac}. 

In this work, we mainly focus on the \emph{high-sparsity regime}, in which \emph{sparse training} methods provide the best known accuracy-vs-sparsity trade-offs.  
We begin by discussing methods for computer vision. Here, 
Gradual Magnitude Pruning (GMP), in which the lowest-magnitude weights are progressively removed throughout training, is a common baseline. In \cite{gale2019state}, GMP was shown to be competitive with more sophisticated pruning methods on image classification models when properly tuned; similar results were later shown for language models \cite{kurtic2022gmp}. 

 The RigL pruning method~\cite{evci2020rigging} is a common, high-performing benchmark for dynamic sparse training. In this method, the weights are initially pruned to the target sparsity and trained through (sparse) stochastic gradient descent. Periodically, however, the mask is updated by selecting weights with the highest magnitude gradient, subject to a limit on the total mask change. The authors run this method using two sparsity targets - Uniform sparsity, where all layers (except the first and last) are pruned to the same proportion, and Erdős–Rényi Kernel (ERK), where layer sparsity targets are set to optimize performance. The authors test their method in the normal-schedule (100 epochs on Imagenet) and 5x training regime, getting results of 73.22\% validation accuracy and 74.63\% validation accuracy at 95\% global (ERK) and uniform sparsity, respectively when training for 500 epochs. Extending training to 10 000 epochs (100x) further allowed the authors to produce 99\% sparse (ERK) ResNet50 models with 68.5\% accuracy on ImageNet. RigL was improved by combining it with ITOP \citep{liu2021atop}, by altering training hyperparameters to encourage mask exploration, which was shown to improve RigL results at medium (80-90\%) sparsity (see Table ~\ref{tab:imagenet_gradual_sota}).
 
 The GraNet\cite{liu2022granet} method extends this approach by making it gradual - either starting from a dense network and performing RigL-like updates while simultaneously increasing sparsity until the target sparsity is achieved, or by starting by a partially sparse (50\%) network and doing the same. Models trained with the sparse-init version of GraNet  achieved 72.3\% validation accuracy at 95\% global sparsity when training for 100 epochs.

 The AC/DC pruning method~\cite{peste2021ac} alternates dense and sparse pruning phases of several epochs each, effectively co-training dense and sparse models. Similar to RigL, AC/DC was tested in the normal and extended training regime, creating 95\% globally sparse ImageNet-1K ResNet50 models with 73.14\% top-1 accuracy, and 68.44\% top-1 accuracy 98\% sparse models after 100 epochs of training. The authors also experiment with extended training times, producing 95\% uniform sparsity ResNet50 models with 74.3\% validation accuracy.
    
     Another successful pruning approach is the combination of Powerpropagation~\cite{schwarz2021powerpropagation} with Top-KAST~\cite{jayakumar2021top}. In Powerpropagation, the weights are reparametrized using $f(w) = w|w|^{\alpha-1}$ for $\alpha > 1$, effectively encouraging high-magnitude weights to continue increasing while lower-magnitude weights are driven toward 0. Top-KAST is a dynamic sparse training scheme that is largely similar to RigL: in Top-KAST, for a target density $D$, the gradients of the top $D'<D$ weights are computed in each backpropagation round and allowed to accumulate, and the masks at these respective sparsities are periodically recomputed. The combination of these two methods results in 77.16\% accuracy at 90\% sparsity when trained for 3x their baseline of 32K steps.

 The recently-proposed ST-3 method~\cite{Vanderschueren2022AreSG} uses the technique of soft thresholding with straight-through gradient estimation to progressively prune neural networks while allowing weights to move more smoothly between the dense and sparse states. Using this method, the authors were able to achieve ImageNet accuracies of between 74\% and 75\% at 96\% sparsity on ResNet-50, depending on the method variant used.

 Additionally, some works have explored the difficulty of sparse optimization~\cite{evci2019difficulty}, explored changes to dense training pipelines to improve sparse training~\citep{tessera2021gradients, jaiswal2022training}, or focused on the creation of sparse accurate neural networks outside of the standard paradigm of simultaneously searching for the optimal mask and weights. Notably, \cite{liu2021atop} explored the impact of mask exploration (that is, the total number of explored parameters at any point in sparse training), demonstrating the positive effect of extended training on both sparse network performance and total number of explored parameters. The STEP~\cite{lu2023step} learning method explored the interaction of sparsity with the Adam optimizer~\cite{kingma2014adam}, finding that the masked weights lead to an incorrect estimate of the second moment during optimization; these observations led to their proposal of a new method for N:M sparsity that alleviates these effects. The GradMax method~\cite{evci2022gradmax} initializes a small neural network, then uses predicted gradients to grow a larger (while still small) neural network by adding additional neurons.

\paragraph{Language models} For language models, the standard compression pipeline consists of two stages: pre-training on a
large unlabeled text corpus followed by fine-tuning on a small and labeled task-specific dataset. The former is used to capture the statistical patterns and relationships that exist in the natural language, allowing the model to recognize and even generate various linguistic patterns. The latter stage, fine-tuning on a downstream task, builds on top of the learned representations and adapts them to solve specific tasks such as text classification, sentiment analysis, duplicate detection, etc. Sparsity has been explored in both stages: pruning during pre-training and pruning during fine-tuning. 

Methods such as Movement Pruning~\cite{sanh2020movement} and The Optimal BERT Surgeon (oBERT)~\cite{kurtic2022optimal} make use of first-order (gradient) and second-order (curvature) information, respectively, to guide pruning decisions during the fine-tuning stage. However, recent work observed two problems with this approach when applied on small datasets: \cite{zhang2022platon} demonstrated instability due to large variability of estimated importance scores, while \cite{huang2021sparse} observed overfitting despite reduced expressive power due to pruning. 
From the practical side, this approach is less favorable for practitioners as it requires extensive pruning-domain knowledge to properly configure pruners for each model and dataset combination. Therefore, the main focus of our work is on the other stage, leveraging already sparse pre-trained models with transfer learning to obtain highly accurate task-specific fine-tuned models. 
 Prune Once for All (Prune OFA)~\cite{zafrir2021prune} and oBERT~\cite{kurtic2022optimal} represent the most recent state-of-the-art techniques addressing this problem. Both methods first prune the model during the pre-training stage, and then apply transfer learning with a fixed sparsity mask to obtain fine-tuned and sparse models on various downstream datasets. 

\paragraph{Impact of sparsification beyond top-1 accuracy} An open area of research is the impact that pruning in general, and the choice of pruning method in particular, have on the resulting model. In particular, pruned models have been shown to be more vulnerable to bias \cite{hooker2019compressed, hooker2020characterising, iofinova2023bias}, and worse at prediction accuracy under distribution shift~\cite{liebenwein_lost_2021}. Recent works by \cite{chen2021evaluating} and \cite{iofinova2023bias} investigate the effects of pruning on a range of model trustworthiness metrics and find mixed results, with sparse neural networks having better calibration, but exaggerating spurious patterns in the existing data. Finally, works such as~\cite{iofinova2022transfer} and ~\cite{chen2021lottery} investigated the capacity of sparse CNNs for domain adaptation via transfer learning, finding that sparsely trained networks can have more generalizable features than dense ones.

\section{The Difficulty of Sparse Pretraining of Vision Models}

\subsection{Background} 

Formally, accurate pruning is a constrained optimization problem which, given the objective of minimizing a loss function $\mathcal{{L}}$, aims to find an ``optimal'' sparsity mask $\mathbf{M^{\star}}$ with a given target sparsity $s$, fraction of zero parameters,\footnote{A \emph{sparsity mask} is simply a binary tensor 
  of the same dimensions as the model, with $0$ at the indices of the sparsified entries, and $1$ at the other indices.} and  weights $\mathbf{W^{\star}}$ such that
\begin{equation}
    \label{eq:pruning}
    \mathbf{M^{\star}, W^{\star}} = \text{argmin}_{\textnormal{mask $\mathbf{M}$}, \textnormal{ weights } \mathbf{W}} \left[ \mathcal{L} (\mathbf{M} \odot  \mathbf{W}) \right] \quad \mathrm{nonzero}(\mathbf{M}) \leq (1 - s) \mathrm{numel}(\mathbf{M}).
\end{equation}

In its general form, where both the optimal mask and the optimal weights must be determined, this approach is NP-complete~\cite{blumensath2008iterative}, even for simple least-squares loss. 
However, this problem can be made tractable if we assume a fixed mask, or we wish to approximate the sparsity of the mask~\cite{axiotis2020sparse}.

In the context of pruning, this procedure can be logically split into 1) determining the  sparsity mask $\mathbf{M}$, which is  often separated from 2) the optimization procedure over the non-zero weights. For instance, the standard Lottery Ticket Hypothesis (LTH) approach~\cite{frankle2018lottery, chen2021lottery} is to first identify a ``ticket'' mask by performing weight selection by magnitude over an already-trained model, followed by SGD-based finetuning, using the initialization and the same set of hyperparameters as for dense training.  

While several novel ways of choosing or updating the sparsity mask choice (step 1), have been investigated, by and large, for the second step, that of optimizing the remaining weights, sparse training methods largely emulate the hyperparameters of the baseline dense model, including the total number of training epochs \cite{gale2019state, jayakumar2020top, evci2020rigging, peste2021ac}. 
However, it is intuitive that the problem of simultaneously finding near-optimal weights and a near-optimal mask may be harder to solve than a standard dense loss minimization problem. 

This naturally motivates an in-depth investigation into the following questions: 
\emph{can optimization over sparse networks converge with the same rate as over dense ones?}, and \emph{are dense training recipes well-suited for sparse training?} 
In this paper, we provide evidence that the answer to both questions is \emph{negative}, suggesting that improved optimizers may be required for obtaining accurate sparse models under reduced training budgets. 

\subsection{Sparse Vision Models Show Evidence of Undertraining}
\label{sec:imagenet_undertraining}

We begin by investigating correlations between the performance and output characteristics of dense and sparse models trained for increasing number of epochs. 
Specifically, we examine three key metrics: \emph{Top-1 accuracy} on the validation/test set, \emph{loss on the train set},  and \emph{prediction entropy} on the validation/test set for the trained models, while scaling the number of training epochs and the associated hyperparameters correspondingly. We detail these metrics below. 

\paragraph{Train Loss and Output Entropy.} We examine model fit to the training data via the training (cross-entropy) loss at the last epoch, and output predictions via the information-theoretic notion of \emph{entropy}. Low prediction entropy implies that the prediction weight is largely concentrated in a single class, while a high entropy suggests that it is spread out over several classes. 
Intuitively, the entropy of the model is related to its ``confidence'' in predictions, and is independent of whether the predictions are correct (and so can be measured on unlabeled data). Conversely, low training loss measures the model's fit to the training data.

We compute the cross-entropy loss and prediction entropy by taking the softmax over the vector of output values of the network and then applying the respective standard formulas, where the cross-entropy is taken with respect to the correct label distribution for the model (1 for the correct class and 0 otherwise). For an output of a network outputting a vector $Z = (z_1, z_2, ..., z_C)$ of size $C$ with correct label $L$, the entropy $H$ and the cross-entropy $CE$ are given by the following formulas:
\begin{equation}
    H(Z) = -\sum \limits_{i=1}^{C} \frac{e^{z_i}}{\sum \limits_{j=1}^{C} e^{z_j}}\log \left(\frac{e^{z_i}}{\sum \limits_{j=1}^{C}e^{z_j}} \right)  \quad \textrm{and} \quad
    CE(Z) = -\log\left(\frac{e^{z_L}}{\sum \limits_{j=1}^{C}e^{z_j}} \right).
\end{equation}
As we will demonstrate, we expect a sufficiently large and well-trained model to have (a) low loss on the training data and (b) fairly low average prediction entropy, while a model that is not well-trained to have high prediction entropy. However, as is conventionally known, continued training on dense and low-sparsity models resulting in overfitting will lower these metrics further.

\subsubsection{Example 1: Sparse Training on ImageNet}

\paragraph{Experimental setup.} 
We first examine validation accuracy on trained sparse and dense ResNet50 models on the ImageNet-1K dataset and compare it to (a) prediction entropy on the validation dataset and (b) the train loss on the last epoch of training.
All models were trained using standard hyperparameters  (see Appendix~\ref{appendix:hyperparameters}) except for the difference
in number of training of epochs in different experiments.
Measurements represent the final accuracy and entropy after the last training epoch, so each marker on the plots represents a full experiment, rather than an intermediate checkpoint. 
Sparse models were pruned with Alternating Compression/Decompression (AC/DC)~\cite{peste2021ac}, likewise adjusting the total number of compressed and decompressed phases to the total run length. AC/DC was chosen as it was among the best-performing methods across all sparsities and training lengths (see Section~\ref{sec:method_selection}). 
We use the FFCV library~\cite{leclerc2022ffcv} for fast loading of the data. In contrast with other runs presented in this paper, we do not use progressive resizing or label smoothing, as the latter explicitly encourages high prediction entropy and cross-entropy. %
In these experiments, we keep the first and last layer dense.%

\begin{figure}[h!]
    \centering
    \begin{subfigure}{0.32\linewidth}
        \includegraphics[width=\linewidth]{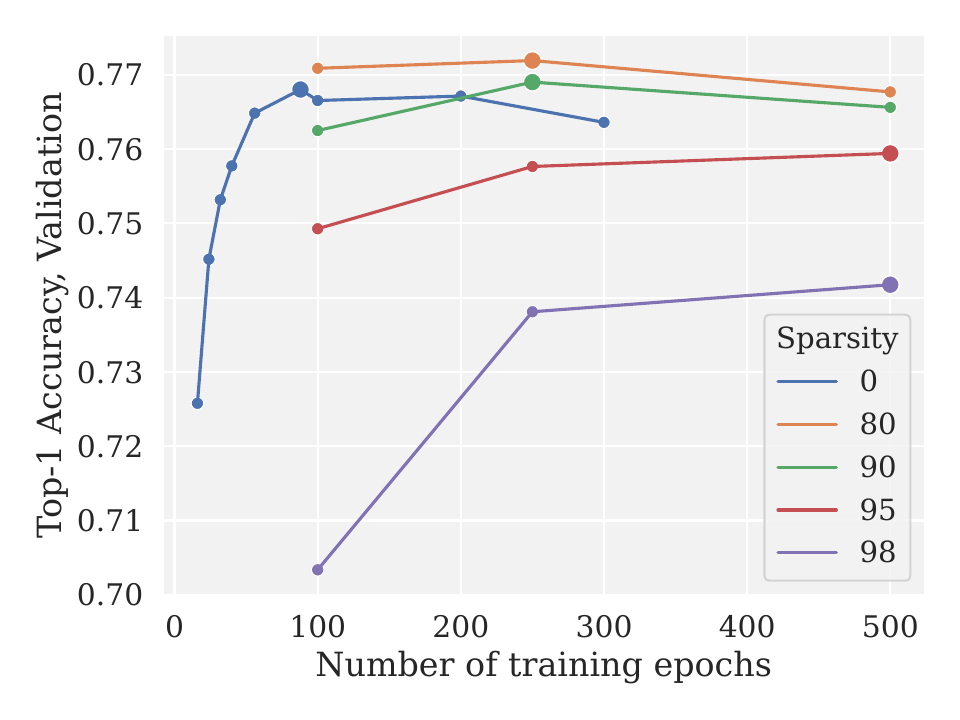}
    \end{subfigure}
    \begin{subfigure}{0.32\linewidth}
        \includegraphics[width=\linewidth]{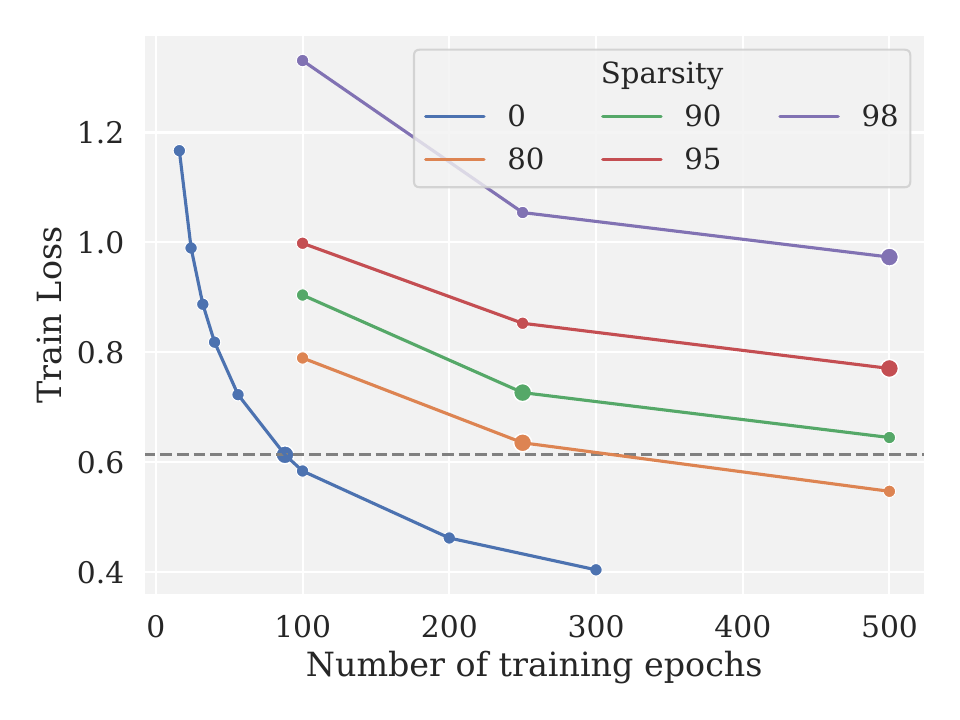}
    \end{subfigure}
       \begin{subfigure}{0.32\linewidth}
        \includegraphics[width=\linewidth]{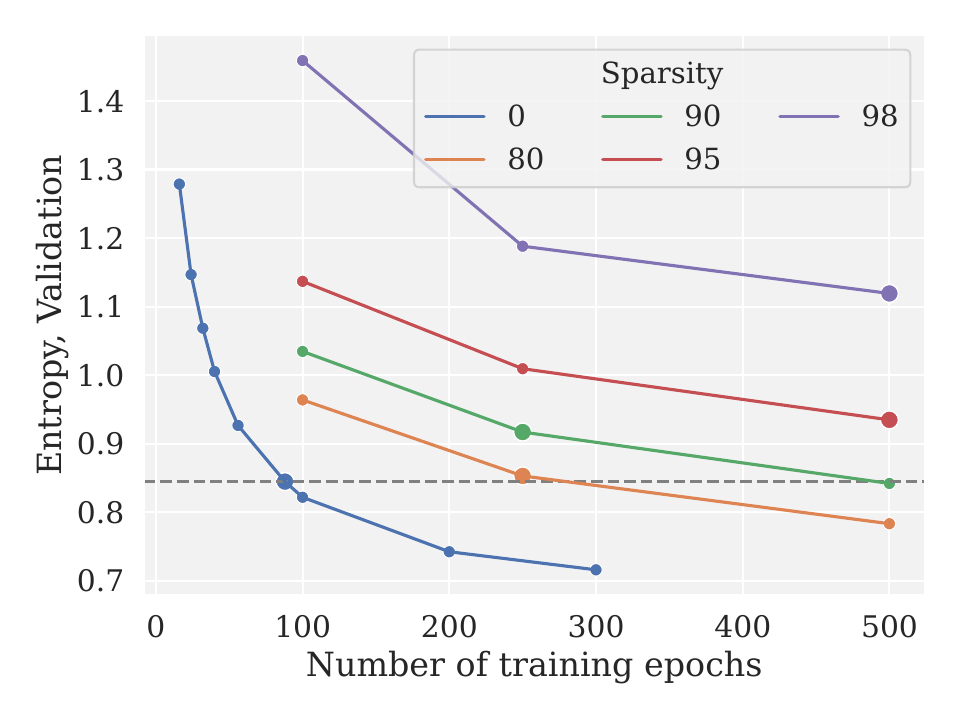}
    \end{subfigure}
    
    \caption{
      Average validation accuracy (left), Train loss at final epoch (center), and Entropy (right) for sparse and dense ImageNet models trained for different numbers of epochs. The highest-accuracy model for each sparsity level is highlighted with a larger marker. The cross-entropy loss and entropy level of the dense model is also shown with a dashed line, to simplify comparison.
    }
    \label{fig:imagenet_acc_uncertainty}
\end{figure}

\paragraph{Results.} Our results are presented in Figure~\ref{fig:imagenet_acc_uncertainty}. On the left panel, we show the top-1 accuracy of the final models. We observe that 80\% and 90\% sparse models reach an accuracy that is similar to dense models, even slightly exceeding dense accuracy at 80\% sparsity. Accuracy does drop at higher  sparsity (95\% and 98\%); this is consistent with the original AC/DC paper and results from other pruning methods. 
Examining accuracy across epoch budgets, and focusing on the best-performing model for each sparsity level, we observe the following:

\begin{itemize} 
\item \emph{The dense model requires the fewest epochs} (88) to reach its best validation accuracy, and extending the training recipe results in \emph{worse performance for the dense model}, commonly known as ``overfitting.'' 
\item \emph{The outcome changes if we examine sparse models}, for which the ideal training length increases with sparsity: 250 epochs for 80\% and 90\% sparse models, and at least 500 epochs---the longest schedule we tried in this experiment---for 95\% and 98\% sparse models. Even at 500 epochs, the accuracy increase/loss decrease for these models does not appear to be saturated. 
\end{itemize}

We now examine loss on the training dataset and prediction entropy for the same experiment in more detail. These two metrics give us two different ways to consider the convergence of our model. The (cross-entropy) loss on the training data shows how well the parameters of the model fit the learning objective; conversely, the prediction entropy on the validation data, although similar in calculation, reflects the model's confidence in its predictions when presented with previously unseen data; additionally, unlike cross-entropy, it is \emph{label-independent}.

We observe that, for all sparsity levels, both metrics behave very similarly. Specifically, both decrease when the number of training epochs is increased, and sparse models trained for the standard 100 epochs show similar training loss and prediction entropy to dense models trained for far fewer epochs. For example, dense models trained for 24 epochs have a similar training loss and prediction entropy to 95\% sparse models trained for 100 epochs, while dense models trained for 100 epochs have a slightly lower training loss and prediction entropy than 80\% sparse models trained for 250 epochs. When we consider the best-performing models at their respective sparsity levels, we find that they have similar training loss and prediction entropy to the top-performing dense model, in cases where such low loss/entropy can be achieved in a reasonable number of epochs (at 80\% and 90\% sparsity); at all sparsities, performance drops for models whose training loss and prediction entropy fall below this value. 

\paragraph{Discussion.} These findings further support our hypothesis that, due to the inherent difficulty of sparse optimization, using standard training recipes is not sufficient for sparse training, 
and suggests that longer training may mitigate this effect. 
Further, results suggest that training loss and/or prediction entropy may be useful criteria to validate that the sparse models are properly trained\footnote{The 98\% sparse model will likely never reach the entropy of the optimal dense model, suggesting that the accuracy may continue to improve with very long training schedules. In fact, the authors of RigL trained a 99\% sparse model for 100 times the dense training time and were not able to saturate its accuracy. See \url{www.github.com/google-research/rigl\#extended-training-results}.}, with the latter criterion being also useful in cases where access to train data, or to any labeled data, is not possible. 

\subsubsection{Example 2: Sparse Training on Celeb-A}
\label{sec:celeba}
To validate our findings, we repeat this experiment on the Celeb-A Dataset~\cite{liu2015deep}. This dataset consists of a combined $202 599$ face images of $10 177$ celebrities collected from the public domain, automatically cropped to the face, and annotated with $40$ binary labels. Due to its content, this dataset is frequently used to study bias in machine learning models, and has also been used in studies on the effect of sparsity on bias~\cite{hooker2019compressed, iofinova2023bias}. 

\begin{figure}[h]
    \centering
    \begin{subfigure}{0.32\linewidth}
        \includegraphics[width=\linewidth]{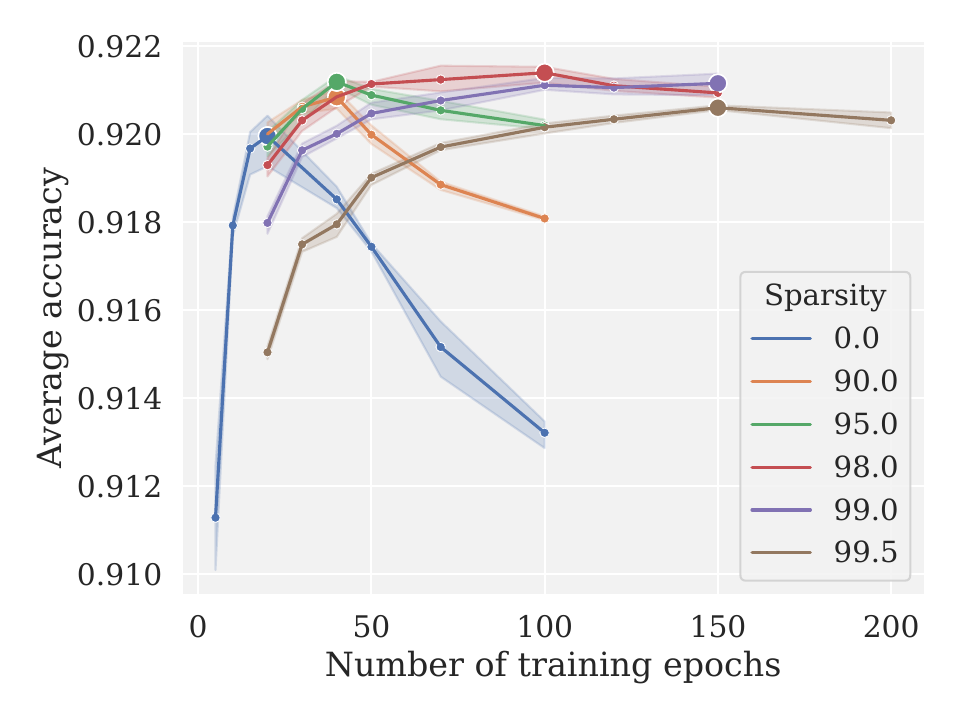}
    \end{subfigure}
    \begin{subfigure}{0.32\linewidth}
        \includegraphics[width=\linewidth]{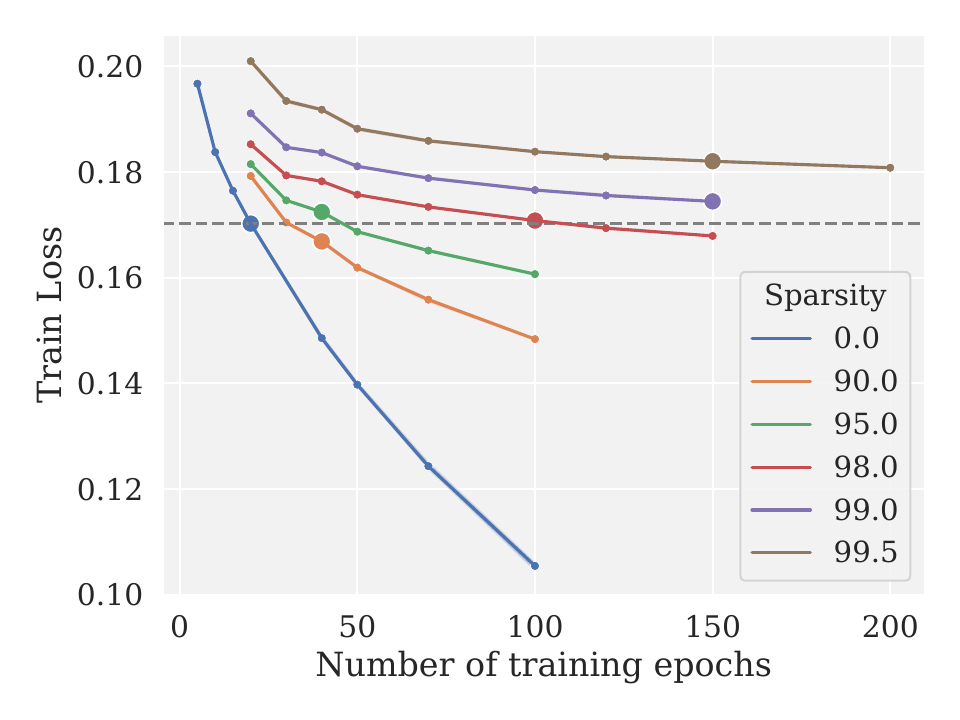}
    \end{subfigure}
       \begin{subfigure}{0.32\linewidth}
        \includegraphics[width=\linewidth]{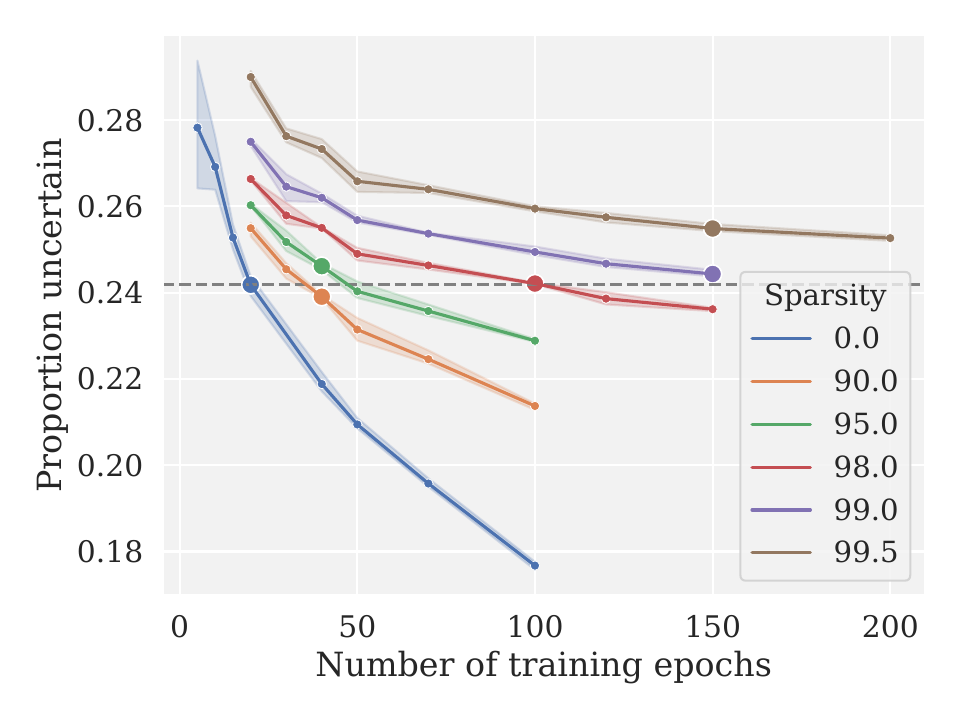}
    \end{subfigure}
    \caption{
      Average validation accuracy (left), train loss at final epoch (center), and uncertainty (right) for sparse and dense CelebA models trained for different numbers of epochs. The highest-accuracy model for each sparsity is highlighted with a larger marker. The cross-entropy loss and entropy level of the dense model is also shown with a dashed line, to simplify comparison.
    }
    \label{fig:celeba_auc_uncertainty}
\end{figure}

    \paragraph{Experimental Setup.} Following \cite{iofinova2023bias}, we train the ResNet18 architecture on this task. We train dense and sparse (90\%, 95\%, 98\%, 99\%, and 99.5\%) models for a varying number of epochs (5-200); in some cases the very low or very high-epoch runs are skipped if it is clear that the duration will not be optimal for that sparsity. Sparse models are produced with a variant of AC/DC, in which the sparsity of the sparse phases ramps up progressively from 90\% to the final target sparsity; this is necessary to prevent layer collapse at very high sparsities. Unlike the ImageNet experiments, here the phase length varies somewhat with duration, due to the extremely short duration of some runs. For each experiment, characterized by a sparsity/epoch-length pair, we measure the accuracy and training loss of the resulting model. In addition, following~\cite{iofinova2023bias}, we measure the \emph{uncertainty}, rather than entropy, of the model predictions on the test set. The prediction uncertainty is computed as follows: first, the sigmoid operator is applied to each logit's output in order to obtain a pseudo-probability of a positive label between 0 and 1; if this quantity is between 0.1 and 0.9, the prediction is considered uncertain. We then compute the proportion of uncertain predictions across the validation dataset.

\paragraph{Results.} The results are presented in Figure~\ref{fig:celeba_auc_uncertainty}. We observe that, consistent with our earlier observations on ImageNet, the optimal training duration goes up with sparsity, with dense models reaching their optimal accuracy at 25 epochs, and 99\% and 95\% sparse models at 150 epochs. Further, we observe that, even as training loss and test uncertainty always decrease with longer training, the overall training loss and proportion of uncertain predictions goes up with sparsity at a fixed training length.
As in the ImageNet example, the highest-performing models at each sparsity have a similar training loss of about 0.17 and mean prediction uncertainty of about 24\%, except for the very sparse 99.5\% model, which has slightly higher 26\% uncertain predictions. 

\paragraph{Discussion.} %
We interpret these results as corroborating evidence that sparse models require longer training compared to dense models
to achieve optimal accuracy before the overfitting starts to take place.

\subsection{State-of-the-Art Accurate Sparse Pre-Training on ImageNet}\label{sec:asp_for_imagenet}

The above observations for vision models suggest that successful sparse training may benefit from an extended training schedule. 
We now build on this idea to achieve state-of-the-art results for the classic ResNet50/ImageNet benchmark by using an extended-training version of AC/DC, which we call AC/DC++. 

\subsubsection{Comparing Sparse Training Methods}%
\label{sec:method_selection}

\begin{figure}[!h]
    \centering
    \begin{subfigure}{0.99\linewidth}
        \includegraphics[width=\linewidth]{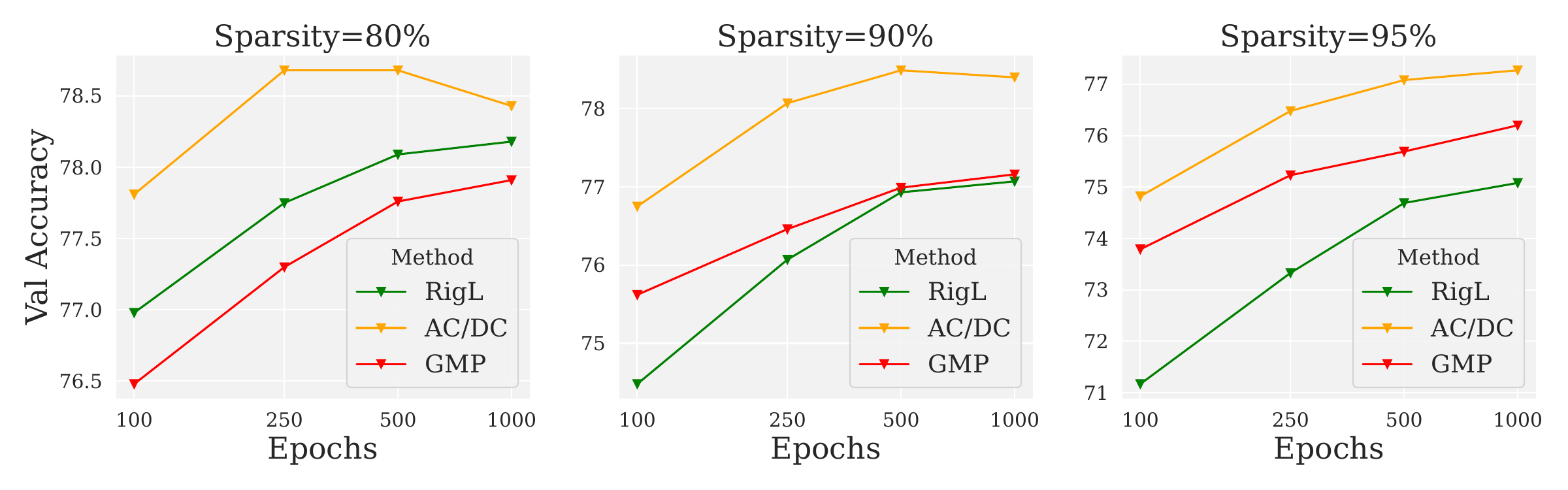}
    \end{subfigure}
    \caption{
       Validation accuracy on ImageNet-1k vs number of epochs for different sparse training methods.
    }
    \label{fig:asp_method_comparison_val_acc}
\end{figure}

\begin{figure}[!h]
    \centering
       \begin{subfigure}{0.99\linewidth}
        \includegraphics[width=\linewidth]{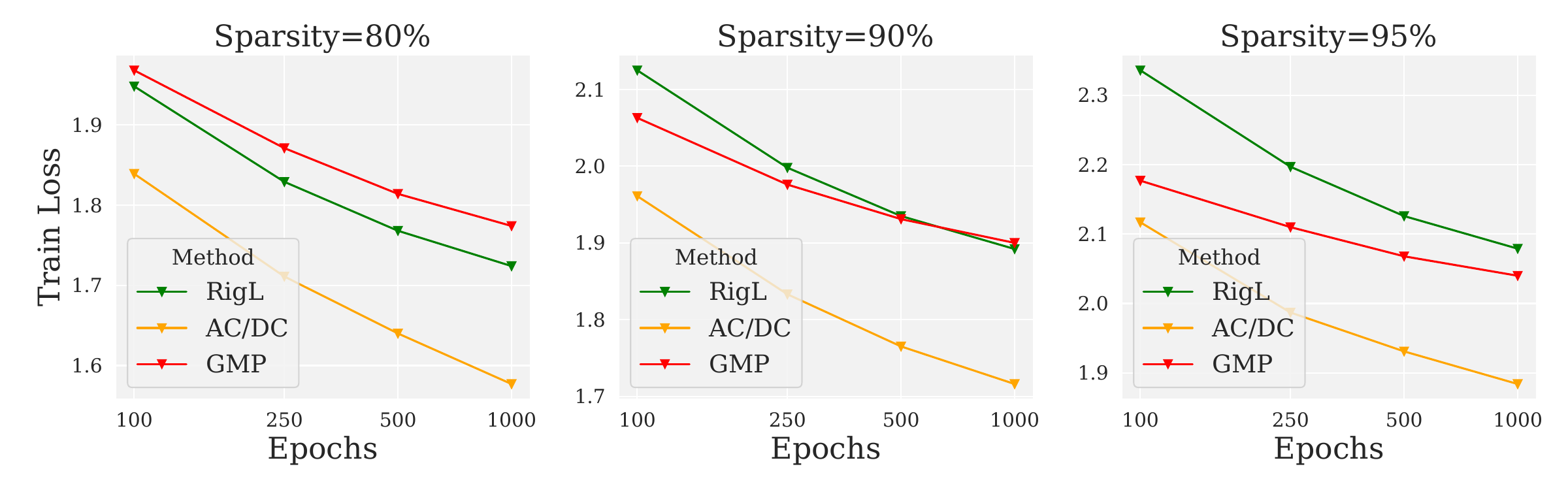}
    \end{subfigure}
    \caption{
       Training loss  on ImageNet-1k vs number of epochs for different sparse training methods.
    }
    \label{fig:asp_method_comparison_train_loss}
\end{figure}

For the following experiments, we start from the current state-of-the-art training approach for ResNet50/ImageNet training, using the Pytorch FFCV package~\cite{leclerc2022ffcv}. In addition to an extended training schedule, we use label smoothing and a linear learning rate decay with warm-up, as well as progressive resizing of input samples \footnote{We follow the setup from the \href{https://github.com/libffcv/ffcv-imagenet}{FFCV ImageNet example} repository for ResNet50.}. 
In this context, we implemented three leading sparse training methods: Gradual Magnitude Pruning (GMP)~\cite{zhu2017prune}, RigL~\cite{evci2020rigging} and AC/DC~\cite{peste2021ac}, which we execute for an increasing number of epochs between 100 (standard) and 1000 (10x). For this, we scale the original training schedule proportionally, following the proportions employed by the original methods. %
For this experiment, models are compressed to 80\%, 90\%, and 95\% sparsity. 
Following the most common experimental setup we prune all weights in convolutional and linear layers (including input convolution and classification head).
The exact training recipe is presented in detail in Appendix~\ref{appendix:hyperparameters}. We note that all the experiments presented in the paper take less than a day on a standard 8-GPU server.
The results, in terms of accuracy and loss vs number of training epochs are presented in Figure~\ref{fig:asp_method_comparison_val_acc}
and Figure~\ref{fig:asp_method_comparison_train_loss}, respectively.  

\paragraph{Results.} 
The results show a strong correlation between how well the methods achieve reduction in loss and their validation accuracy. This reinforces the point that sparse training methods saturate slower, both in terms of training loss and validation accuracy. 
This has also been investigated by prior work: Gale et al.~\cite{gale2019state} found that extended training did improved results for GMP in some cases, while RigL~\cite{evci2020rigging} and Powerpropagation~\cite{schwarz2021powerpropagation} found diminishing improvements. 
At the same time, we notice a significant difference between methods: specifically, 
AC/DC starts at a slightly better accuracy point, and consistently outperforms other methods both in terms of loss achieved, and in terms of validation accuracy, as we increase training time. (This is consistent with the AC/DC original results, executed at 100 epochs~\cite{peste2021ac}.)   
We observe that this correlates with the theoretical computational cost (FLOPs) of the methods: 
AC/DC will use more FLOPs than other methods due to the dense training phases, while GMP uses more FLOPs than RigL due to gradually increasing sparsity. 
In turn, this could also be correlated with the amount of mask exploration performed by the algorithm during training. 
At low sparsity RigL performs slightly better than GMP, but for higher sparsity GMP appears to perform better. 
For the smallest 80\%, 90\% AC/DC reaches a saturation point, whereas in all other setups model performance continues to improve with training budget.

\subsubsection{Sparsity-vs-Accuracy Results}
\label{sec:imagenet_results}
\begin{figure}[h]
    \centering
    \begin{subfigure}{0.49\linewidth}
        \includegraphics[width=\linewidth]{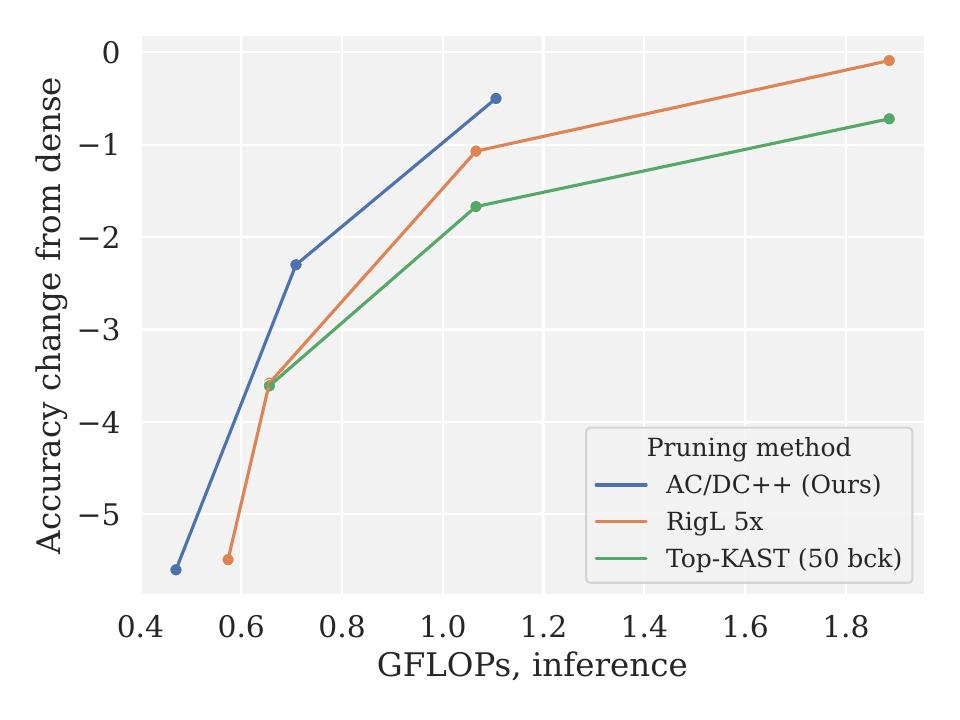}
    \end{subfigure}
       \begin{subfigure}{0.49\linewidth}
        \includegraphics[width=\linewidth]{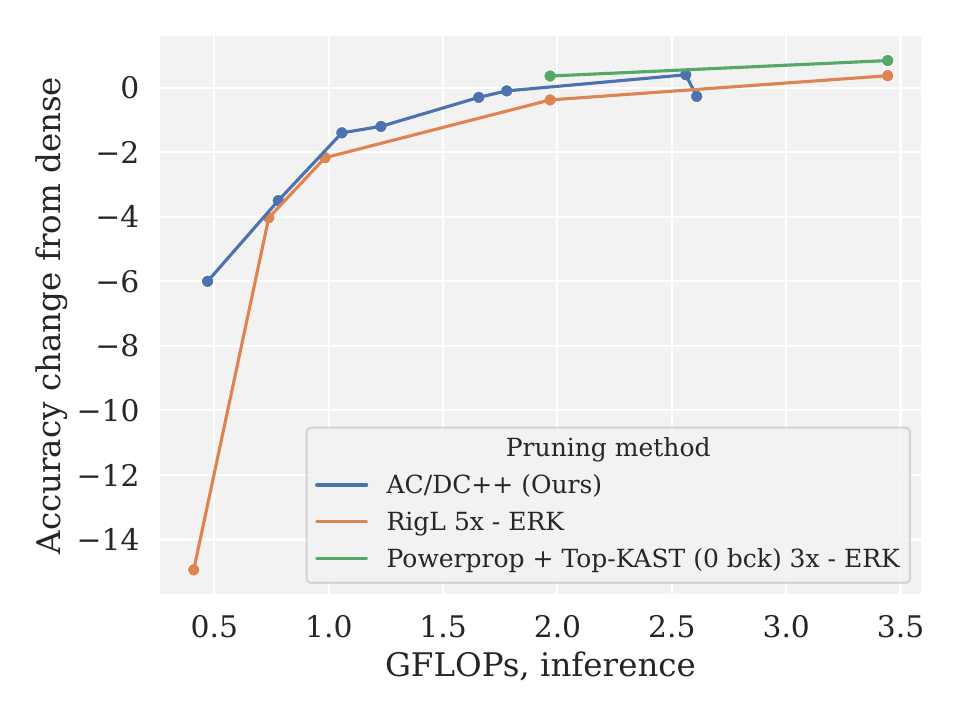}
    \end{subfigure}    
    \caption{Comparison of
    Accuracy change from dense baseline as a function of Inference FLOPs for leading sparse training methods, under uniform sparsity constraints (\textbf{left}) and global sparsity constraints (\textbf{right}). Due to a lack of a standard benchmark, global and Erdős–Rényi Kernel (ERK) sparsity constraints were grouped together. Both sparsity schedules of AC/DC++ (with all layers sparsified and with the first and last layer kept dense) are plotted together.
    }
    \label{fig:imagenet_method_comparison}
\end{figure}

\begin{table}[t!] \centering
	\centering
	
	\resizebox{0.75\textwidth}{!}{
		\begin{tabular}{@{}lcccccc@{}}
			\toprule
			\multirow{1}{*}{} & \multicolumn{2}{c}{Top-1 accuracy (\%)} &$\Delta$ Accuracy  & \multicolumn{1}{c}{Sparsity} & \multirow{1}{*}{Remaining   } & \multicolumn{1}{c}{Inference FLOPs} \\
			\cmidrule(l{3pt}r{3pt}){2-3}
			\multirow{1}{*}{Method} & Dense {\small($D$)}  & Sparse  {\small($S$)}& {\small${100 \times \frac{ (S-D)}{D}}$} &   (\%) & \# of params & prop. of dense\\
			\midrule
                Sparse Training \\
                AC/DC \cite{peste2021ac} & 76.8 & 75.03 & -1.77 & 90 & 2.56 M & 0.18 \\
                GraNet($s_0=0.5$) \cite{liu2022granet} & 76.80 & 74.5 & -1.3 & 90 & - & 0.20 \\
                Powerpropagation + Top-KAST FLD \cite{schwarz2021powerpropagation} & 76.8 & 75.23 & -1.57 & 90 & - & - \\
                Powerpropagation + Top-KAST ERK \cite{schwarz2021powerpropagation}& 76.80 & 75.74 & -1.06 & 90 & - & 0.24 \\
			RIGL ERK 1x \cite{evci2020rigging} & 76.80 & 73.00 & -4.94 & 90 & - & 0.24 \\
                RIGL-ITOP ERK 1x \cite{liu2021atop} & 76.80 & 73.82 & -2.98 & 90 & - & 0.24 \\
                ST-3 \cite{Vanderschueren2022AreSG} & 77.10 & 75.28 & -1.82 & 90 & - & 0.24 \\
			        STR \cite{kusupati2020soft} & 77.01 & 74.31 & -3.51 & 90.23 & 2.49 M & -\\
                Variational Dropout \cite{molchanov2017variational} & 76.69 & 73.84 & -3.72 & 90.27 & 2.49 M & -   \\
   \midrule
                Post-training sparsification \\
                Global Magnitude \cite{singh2020woodfisher} & 77.01 & 75.15 & -2.42 & 90 & 2.56 M & - \\
			WoodFisher \cite{singh2020woodfisher} & 77.01 & 75.21 & -2.34  & 90 & 2.56 M & - \\
   \midrule
   
   Extended sparse training \\

                AC/DC++ 5x (this work) & 78.78 & 78.49 &  -0.29 & 90 & 2.60 M  & 0.2\\
                AC/DC++ FLD 5x (this work) & 78.78 & 78.6 & -0.18 & 90 & 4.45 M & 0.22 \\
                GMP FLD 1.5x \cite{gale2019state} & 76.69 & 75.16  & -1.53 & 90 & - & -  \\
                GraNet($s_0=0.5$) 2.5x \cite{liu2022granet} & 76.80 & 76.4 & -0.4 & 90 & - & 0.20 \\
                Powerpropagation+Top-KAST ERK 3x\cite{schwarz2021powerpropagation} & 76.80 & 77.16 & +0.36 & 90 & - & 0.24 \\
                RIGL ERK 5x \cite{evci2020rigging} & 76.80 & 76.42 & -0.38 & 90 & - & 0.24 \\
                RIGL-ITOP ERK 5x \cite{liu2021atop} & 76.80 & 75.50 & -1.30 & 90 & - & 0.24 \\
                \midrule
			\midrule
   
Sparse Training \\
                AC/DC \cite{peste2021ac} & 76.8 & 73.14 & -3.66 & 95 & 1.28 M & 0.11\\
			
			GraNet($s_0=0.5$) \cite{liu2022granet} & 76.80 & 72.3 & -6.5 & 95 & - & 0.12 \\
                
                Powerpropagation + Top-KAST FLD\cite{schwarz2021powerpropagation} & 76.8 & 73.25 & -3.55 & 95 & - & - \\
			RIGL ERK 1x \cite{evci2020rigging} & 76.80 & 70.00 & -8.85 & 95 & - & 0.12 \\
                ST-3 \cite{Vanderschueren2022AreSG} & 77.10 & 74.46 & -2.64 & 95 & - & 0.13 \\
                STR \cite{kusupati2020soft} & 77.01 & 70.40 & -8.58  & 95.03 & 1.27 M & - \\
                Variational Dropout \cite{molchanov2017variational} & 76.69 & 71.81 & -6.36 & 94.94 & 1.30 M & -  \\
   \midrule
                Post-training sparsification \\
                Global Magnitude \cite{singh2020woodfisher} & 77.01 & 71.72 & -6.29  & 95 & 1.28 M & - \\
			WoodFisher \cite{singh2020woodfisher} & 77.01 & 72.12 & -6.89 & 95 & 1.28 M & - \\
                M-FAC \cite{frantar2021m} & 77.01 & 72.6 & -4.41 & 95 & 1.28 M & - \\
   \midrule	
   Extended sparse training \\
            
            AC/DC++ 10x (this work) & 78.78 & 77.27 &  -1.48 & 95 & 1.33 M & 0.13\\	
            AC/DC++ FLD 10x (this work) & 78.78 & 77.7 & -1.08 & 95 & 3.28 M & 0.14\\
            GMP FLD 1.5x  \cite{gale2019state} & 76.69 & 72.71  & -3.98 & 95 & 1.28 M & - \\
            RIGL ERK 5x \cite{evci2020rigging} & 76.80 & 	74.63 & -2.17 & 95 & 1.28 M & 0.12\\
        \midrule
        \midrule
            Sparse training \\
            AC/DC \cite{peste2021ac} & 76.8 & 68.44 & -9.36 & 98 & 0.7 M & 0.06 \\
            ST-3 \cite{Vanderschueren2022AreSG} & 77.10 & 70.46 & -6.64 & 98 & - & 0.07 \\
            STR \cite{kusupati2020soft} & 77.01 & 70.40 & -8.58  & 98 & - & - \\
            Variational Dropout \cite{molchanov2017variational} & 76.69 & 64.52 & -15.87 & 98.57 & 0.36 M & -   \\
            \midrule
            Post-training sparsification \\
            M-FAC \cite{frantar2021m} & 77.01 & 67.5 & -9.51 & 98 & - & - \\
            WoodFisher \cite{singh2020woodfisher} & 77.01 & 65.55 & -11.46  & 98 & 0.51M & - \\
            \midrule
            Extended sparse training \\
            AC/DC++ 10x (this work) & 78.78 & 74.06 &  -4.72 & 98 & 0.51 M  & - \\
            AC/DC++ FLD 10x (this work) & 78.78 & 76.6 & -2.28 & 98 & 2.58 M & 0.09 \\
            \midrule
            \midrule
            Sparse training \\
            ST-3 \cite{Vanderschueren2022AreSG} & 77.10 & 63.88 & -13.22 & 99 & - & 0.04 \\
            \midrule
            Extended sparse training \\
            AC/DC++ FLD 10x (this work) & 78.78 & 72.7 & -6.08 & 99 & 2.34 M & 0.06 \\
            RIGL ERK 5x \cite{evci2020rigging} & 76.80 & 	61.86 & -15.94 & 99 & - & 0.05 \\
            RIGL ERK 10x \cite{evci2020rigging} & 76.80 & 	63.89 & -12.91 & 99 & - & 0.05 \\
            RIGL ERK 50x \cite{evci2020rigging} & 76.80 & 	66.94 & -9.86 & 99 & - & 0.05 \\
            RIGL ERK 100x \cite{evci2020rigging} & 76.80 & 	68.15 & -8.65 & 99 & - & 0.05 \\
            
			\bottomrule
		\end{tabular}
	}
        \vspace{5pt}
	\caption{Comparison between modern sparse training methods on ImageNet-1k with ResNet-50 models for various sparsity targets. ERK refers to the Erdos-Renyi Kernel sparsity distribution. FLD refers to the first and last layers being dense (AC/DC++) or the first layer being dense and the last layer being 80\% sparse (GMP, PowerPropagation).}
 \label{tab:imagenet_gradual_sota}
\end{table}

\paragraph{Goals and Metrics.} 
Based on these results, in this section, we aim to improve the best known sparsity-versus-accuracy trade-offs by performing a thorough ablation over sparsities and training length parameters.  
We compare our results to the highest-performing previously published sparse training methods. In particular, we compare an extended-training version of AC/DC, which we call AC/DC++, results reported in the original RigL, ST-3, and Powerpropagation papers, as well as many other existing pruning methods.\footnote{The most successful Powerpropagation approach presented in the paper combines this method with Top-KAST; we use this benchmark, as it performs better than Top-KAST individually}. All methods are described in Section~\ref{sec:related_work}. In cases where the authors conducted extended training using their method, we present those numbers, and we use the FLOPs-optimized ST-3$^{\sigma}$ variant. AC/DC++ candidate models were trained for four preset training lengths (1x, 2.5x, 5x and 10x the standard ImageNet training time on ResNet50) at all sparsity levels, and we chose the best results obtained by ablating over the length of the training run. 

As different methods have different computational budgets and different dense baselines, to ensure a fair comparison, we examine the model performance both in terms of \emph{Top-1 Validation accuracy}, and the \emph{Top-1 Validation accuracy difference from the corresponding dense baseline}. We use the best available numbers originally reported in the papers introducing the methods for comparisons.

\paragraph{Experimental Setup.} 
We compare two pruning regimes. First, we consider \emph{Uniform Pruning}, in which every layer is pruned exactly to the target sparsity, except for the first and last layer, which are left dense. Second, we consider the \emph{Global/Nonuniform Pruning} regime, in which the sparsity budget is set globally. Different works apportion the global budget differently, and also differ with respect to which parts of the network are subject to the global constraint. In particular, Extended GMP \cite{gale2019state} and Top-KAST do not prune the first layer, prune the last layer to a fixed 80\% sparsity, and prune the other layers using a global magnitude criterion. RigL uses an Erdős–Rényi-Kernel distribution for layer sparsity targets, and leaves only the first layer dense. The original AC/DC work uses global sparsity and prunes all convolutional and FC layers. Therefore, to create a more fair comparison, we consider estimated Floating-Point Operations (FLOPs) necessary for inference; these are computed as in \cite{evci2020rigging}. Using FLOPs also equalizes methods across slight variations in ResNet50 architectures, and so we use it also for the Uniform pruning comparison. In addition, we use two pruning schedules for AC/DC++: one which leaves the first and last layer dense and prunes the remaining layers using a global magnitude criterion, and one that prunes all layers using the global magnitude criterion. We do not ablate between the two, but rather present both sets of results in Figure~\ref{fig:imagenet_method_comparison} (jointly) and Table~\ref{tab:imagenet_gradual_sota} (separately). 

We emphasize two key points regarding our comparisons: 
\begin{enumerate}
\item Looking at accuracy alone favors  AC/DC++, as it has a higher dense baseline: since we use several recent training innovations, the dense model can reach close to 79\% dense accuracy over 100 epochs. %
Therefore, it becomes more challenging to maintain the performance of the dense model for highly sparse model
compared to less-optimized baseline. 
\item This is why we also examine \emph{accuracy difference relative to the dense baseline}: this favors other methods, as they are benchmarked against a standard-recipe model that reaches lower 76.8\% accuracy (77.1\% for ST-3).
\end{enumerate}

\paragraph{Results.} The results are presented in Figure~\ref{fig:imagenet_method_comparison} and Table~\ref{tab:imagenet_gradual_sota}. We observe that, for uniform pruning budgets, the AC/DC++ models outperform other methods, both in terms of absolute and relative validation accuracy. This is true even when we consider extended-training schedules for other methods, although we believe we are the first to systematically investigate the impact of increasing training schedules at these sparsity levels.\footnote{In prior work, RigL executed >5x extended training for a 99\%-sparse model only~\cite{evci2020rigging}.} When looking at models trained with global pruning budgets, we observe that AC/DC++ obtains the highest absolute validation accuracy, compared to results reported previously in literature. When considering accuracy change from the dense line, AC-DC++ loses less accuracy than other methods at very high sparsities (lowest FLOPs), despite having the highest-performing dense baseline; at lower sparsity (90\%), it is competitive with other extended training methods. 

\subsection{Additional Results}

\paragraph{Different sparsity patterns}
Since the ResNet models increase the number of channels with the decrease of feature map resolution, one
would expect that bottom layers (those with more channels) should be pruned more aggressively compared to 
the one with less channels. Here our results are consistent with the known observations from literature 
that global sparsity achieves higher performance for the same sparsity. In addition, we have carried out experiments
with the more hardware-friendly block-4 sparsity pattern. 

\begin{figure}[!h]
    \centering
    \begin{subfigure}{0.49\linewidth}
        \includegraphics[width=\linewidth]{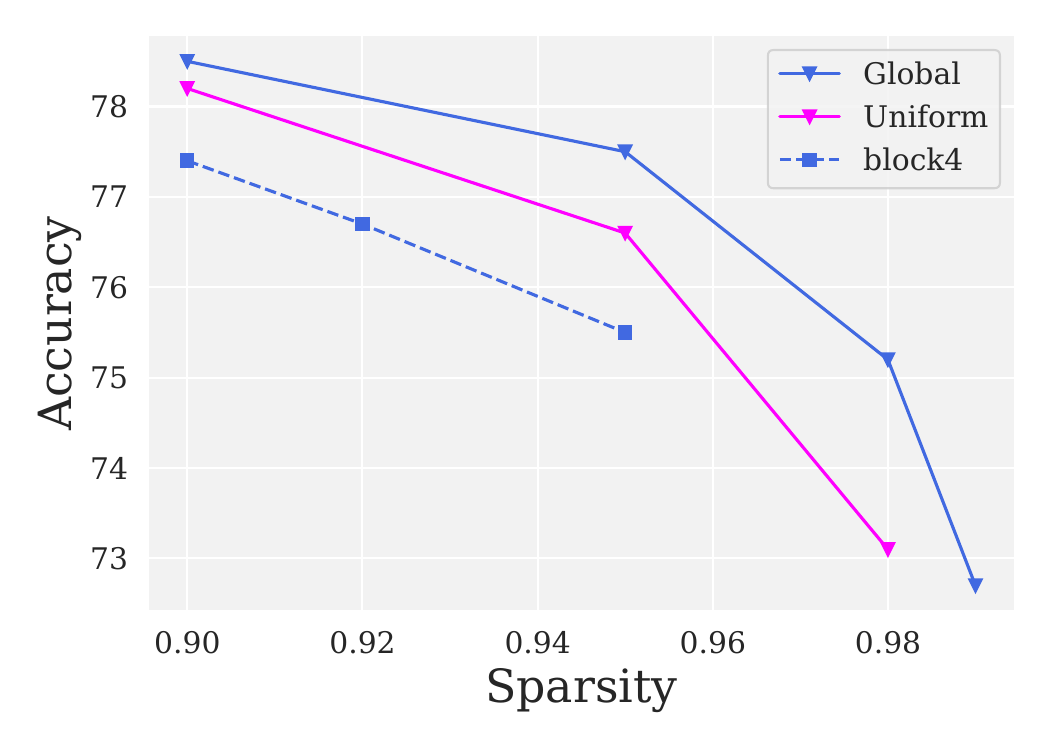}
    \end{subfigure}
    \caption{
        Accuracy vs sparsity for different sparsity distributions.
        Block4 denotes global pruning with weights pruned in groups of 4. 
    }
    \label{fig:acc_global_vs_uniform}
\end{figure}

\paragraph{MobileNet results}
In addition, we conducted sparsification of the MobileNet-V1 model \cite{howard2017mobilenets}, a
CNN optimized for inference on mobile devices. We applied AC/DC for 1000 epochs with sparsity targets 75\% and 90\%
using a similar training recipe to ResNet50 except for some differences specified in Appendix \ref{appendix:hyperparameters}.
To achieve the best results we do not prune the input convolution, as well as the classification head and depthwise convolutions,
due to their minor contribution to the overall amount of FLOPs and significant impact on the performance of the model.

\begin{table}[t!] \centering
	\centering

	\resizebox{0.5\textwidth}{!}{
		\begin{tabular}{@{}lcccc@{}}
			\toprule
			\multirow{1}{*}{} & \multicolumn{2}{c}{Top-1 accuracy (\%)} &Relative Drop  & \multicolumn{1}{c}{Sparsity}  \\
			\cmidrule(l{3pt}r{3pt}){2-3}
			\multirow{1}{*}{Method} & Dense {\small($D$)}  & Pruned  {\small($P$)}& {\small${100 \times \frac{ (P-D)}{D}}$} \\
			\midrule
            \multirow{2}{*}{AC/DC++} & 72.74 & 72.49 & -0.25 & 75.00 \\	
			  \cmidrule{2-5}
             & 72.74 & 70.80 &  -1.94 & 90.00 \\	
			\bottomrule
		\end{tabular}
	}
        \vspace{5pt}
	\caption{Sparse training of MobileNet-V1 with AC/DC++ on ImageNet-1k.}\label{tab:mobilenet_pruning}
\end{table}

With a longer training recipe one can achieve almost negligible accuracy drop at 75\% sparsity
and moderate performance decrease at 90\%. The results can be found in Table~\ref{tab:mobilenet_pruning}.

\subsection{Mask analysis}

Sparse methods considered in our work differ in the amount of sparsity mask exploration. 
GMP gradually increases sparsity; once the weight is pruned, it is never reintroduced.
RigL decreases the fraction of updated parameters following the cosine annealing rule:
\begin{equation}
f_{decay} (t; \alpha; T_{end}) = \frac{\alpha}{2} \left(1 + \cos \left(\frac{\pi t}{T_{end}}\right)\right)
\end{equation}
This fraction of connections is dropped and reintroduced in a single step.
AC/DC makes all parameters trainable on decompression phases, therefore any parameter could 
be potentially reintroduced. However, as shown later, some fraction of weights 
remains zero even on decompression. To measure the difference between two consecutive sparsity masks
we compute their IoU (Intersection over Union), the amount of parameters that are nonzero for both checkpoints
divided by the amount of parameters that are nonzero in either of checkpoints. High IoU value (close to 1) 
means that two sparsity masks overlap significantly, whereas low IoU (close to 0) implies low similarly between
sparsity masks.

\begin{figure}[!h]
    \centering
    \begin{subfigure}{0.49\linewidth}
        \includegraphics[width=\linewidth]{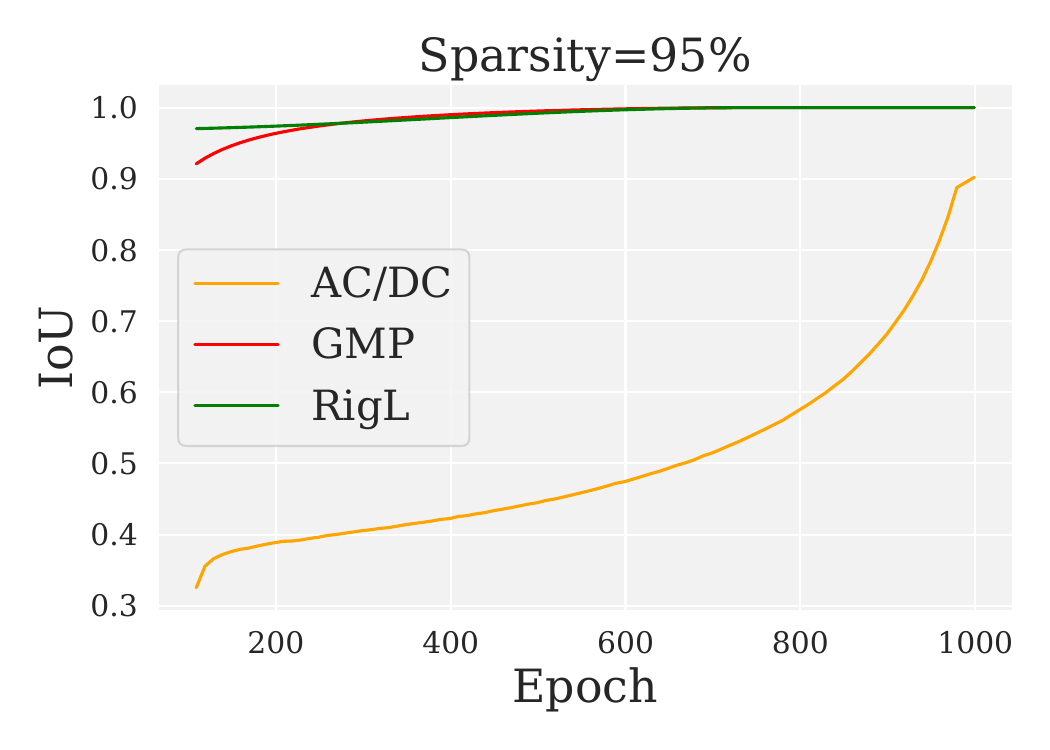}
    \end{subfigure}
    \begin{subfigure}{0.49\linewidth}
        \includegraphics[width=\linewidth]{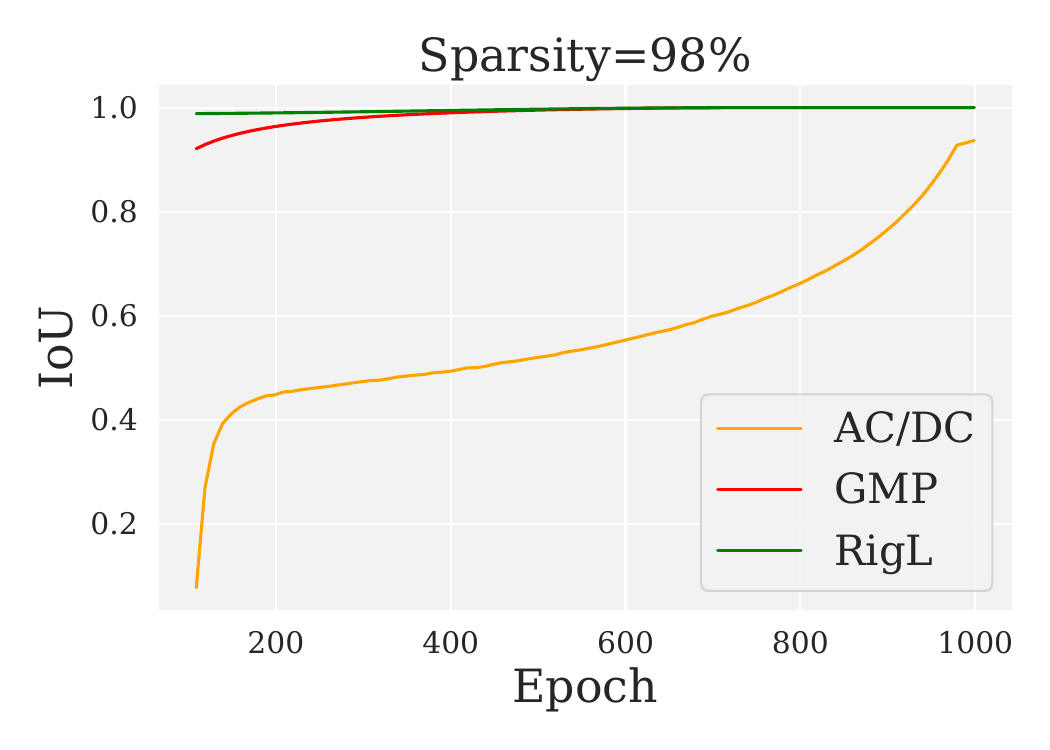}
    \end{subfigure}
    \caption{Mask IoU between two consecutive checkpoints.}
    \label{fig:sparsity_mask_iou}
\end{figure}

We have taken checkpoints saved on the $109^{th}, 119^{th}, \ldots 999^{th}$ epoch (taken at the end of every AC/DC step and at the same epochs for other methods for a consistent comparison) collected during 1000 epochs runs 
with 95\% and 98\% target sparsity and measured 
IoU between two consecutive masks for parameters being pruned (skipping biases and batch norm parameters). 
For GMP and RigL, mask IoU can be computed analytically based on the update rule. The evolution of sparsity mask IoU during training is presented on Figure~\ref{fig:sparsity_mask_iou}. One can observe that AC/DC shows significantly stronger mask exploration compared to GMP and RigL.
This behavior could account for the better performance of AC/DC as a sparse trainer.  

\subsection{Structured sparsity in unstructured sparse models}
\label{sec:channel_sparsity}

In this work, we consider only \emph{unstructured} sparsity, therefore groups of weights and entire channels in particular 
do not have to be sparse. However, we observed that some of the channels in convolutional kernels are entirely pruned.
The effect becomes more pronounced with higher sparsity and longer training. 

In Table \ref{tab:channel_sparsity}, we show the fraction of sparse \emph{output} channels for layers
in the front, middle and the end of the model and the global fraction of zero channels across all convolutional layers 
in the model. 
We observe that channel sparsity increases proportionally with the unstructured sparsity target and with training time, and that, for high sparsity, we obtain a very high proportion of zeroed-out output channels, especially in the wider bottom layers. This is in tune with previous work observing the emergence of structured sparsity in dynamic sparse training~\cite{peste2021ac, iofinova2022transfer, yin2023dynamic}. We provide a first explanation for this behavior in the next section. 

\begin{table}[t!] \centering
	\centering
	\resizebox{0.5\textwidth}{!}{
		\begin{tabular}{cccccc}
			\toprule
            \multirow{2}{*}{} & \multirow{1}{*}{} & \multicolumn{4}{c}{Channel sparsity} \\
			Sparsity & Epochs & \texttt{layer1.0.conv2} &  \texttt{layer2.1.conv2} & \texttt{layer4.2.conv2} & avg  \\
			\midrule
            \multirow{2}{*}{80}	& 100 & 0 & 0 & 0 & 1.07 \\
            	& 1000 & 18.75	& 0	& 35.35	& 4.91 \\
            \midrule
            \multirow{2}{*}{90}	& 100 & 4.69 & 1.56 & 0.78 & 3.58 \\
            	& 1000 & 31.25 & 19.53 & 61.91 & 10.68 \\
            \midrule
            \multirow{2}{*}{95}	& 100 & 0 & 7.03 & 22.85 & 8.8 \\
            	& 1000 & 26.56& 22.66 & 80.27 & 16.76 \\
            \midrule
            \multirow{2}{*}{98}	& 100 & 7.81 & 12.5	& 75.59	& 23.65 \\
            	& 1000 & 48.44 & 37.5 & 96.68 & 27.93 \\
			\bottomrule
		\end{tabular}
	}
        \vspace{5pt}
	\caption{Fraction of zero output channels for specific layers and on average. }\label{tab:channel_sparsity}
\end{table}

\subsection{Impact of weight decay on sparsity and model performance}

Recall that AC/DC makes all parameters trainable on decompression, therefore, one might expect
that sparsity on decompression phase would be near zero. However, we observed that a large fraction of weights remains zero even when the sparsity mask is not imposed. This effect is linked to the channel sparsity discussed in the previous section: once a channel is completely zeroed out, it will continue to receive zero gradients even when the sparsity mask is removed. Further, we provide evidence that this phenomenon is linked to increasing the weight decay mechanism, and in particular the value of this parameter: intuitively, weight decay slowly drives weights towards zero; whenever a full channel is zeroed out in the compression phase, it remains ``captured'' under the sparsity mask until the end of training.

We investigated this empirically by training ImageNet models on ResNet50 with 95\% compression sparsity using AC/DC++ for 100 epochs, varying the weight decay parameter from $10^{-6}$ to $10^{-3}$. Our results are shown in Figure~\ref{fig:sparsity_for_weight_decay}. We observe that the fraction of zero parameters increases with the magnitude of weight decay and also over the course of training. Concretely, we observe that all weight decay values lead to almost fully dense models during the first decompression phase. From there, very low weight decay values of $10^{-6}$ and $10^{-5}$ lead to very little sparsity during the next two decompression phases, and about 10\% sparsity during the final five. Conversely, very high weight decay of $10^{-3}$ leads to an immediate jump to 50\% sparsity during the second decompression phases, which then increases to 60\% over the rest of training. The intermediate value of $10^{-4}$, which is the standard setting and was used in our experiments, leads to an intermediate sparsity, which gradually rises to about 24\% over successive decompression phases. 

We further present the accuracy of the resulting models in Table \ref{tab:accuracy_for_weight_decay}. We observe that properly setting the weight decay hyperparameter is crucial for good performance of AC/DC++, and confirm that the standard value $10^{-4}$ is close to the optimal value, as the Table \ref{tab:accuracy_for_weight_decay} shows.

\begin{center}
    \begin{minipage}{0.49\linewidth}
    \includegraphics[width=\linewidth]{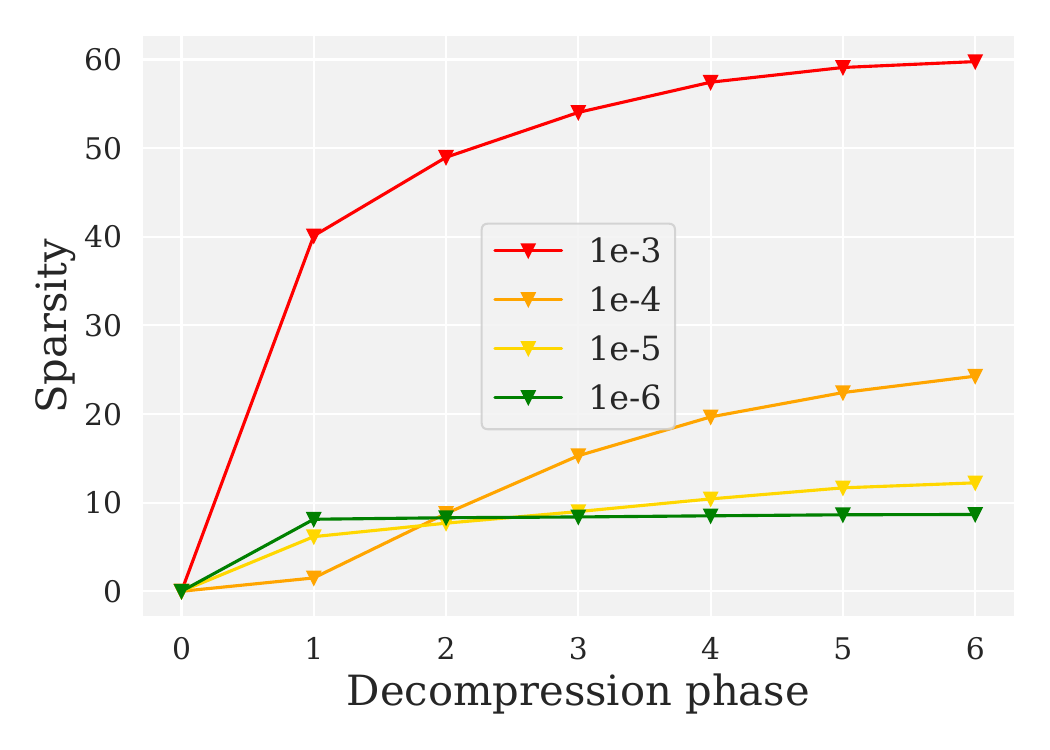}
    \captionof{figure}{Sparsity on decompression phases for 100-epoch AC/DC++ runs with varying values of weight decay. We point out that on decompression phases no sparsity is enforced. }
    \label{fig:sparsity_for_weight_decay}
    \end{minipage}
    \begin{minipage}{0.49\linewidth}
    \centering
    \begin{tabular}{cc}
    \toprule
    Weight decay & Top-1 accuracy (\%) \\
    \midrule
    $10^{-6}$ & 70.52 \\
    $10^{-5}$ & 73.54 \\
    $10^{-4}$ & 74.90 \\
    $10^{-3}$ & 68.57 \\
    \bottomrule
	\end{tabular}
    \captionof{table}{Accuracy vs weight decay.}
    \label{tab:accuracy_for_weight_decay}
    \vspace{20pt}
    \begin{tabular}{cc}
    \toprule
    Decompression sparsity & Top-1 accuracy (\%) \\
    \midrule
    0	& 74.82 \\
    50	& 75.09 \\
    60	& 74.81 \\
    70	& 74.83 \\
    80	& 74.38 \\
    \bottomrule
	\end{tabular}
    \captionof{table}{Accuracy vs decompression sparsity.}
    \label{tab:accuracy_vs_decompression_sparsity}
    \end{minipage}
\end{center}

\paragraph{Sparse Decompression.}
Building on this observation, we ask 
how imposing a fixed minimal sparsity also on \emph{decompression stage} impacts the final performance.
We conducted a few 100-epoch long AC/DC experiments with 95\% target sparsity and set the sparsity on decompression
to some fixed value, smaller than compression sparsity. We observe that the performance is almost unaffected 
up to 80\% decompression sparsity, showing that full mask exploration is not necessary during training. 

\subsection{Loss landscape analysis}

In order to get more insights into the reasons for the difficulty of optimizing sparse neural networks, we investigate two properties of the loss landscape. 
First, we measured \emph{landscape sharpness at the end of the training}, defined as the maximal eigenvalue of the Hessian matrix, for all sparse training methods considered, various sparsities
and number of training epochs and compared with the one of standard dense training. 
Second, we interpolated the training and validation loss between 
checkpoints obtained at intermediate steps throughout the 1000 epoch AC/DC run with 95\% target sparsity. 

\begin{figure}[!h]
    \centering
       \begin{subfigure}{0.99\linewidth}
        \includegraphics[width=\linewidth]{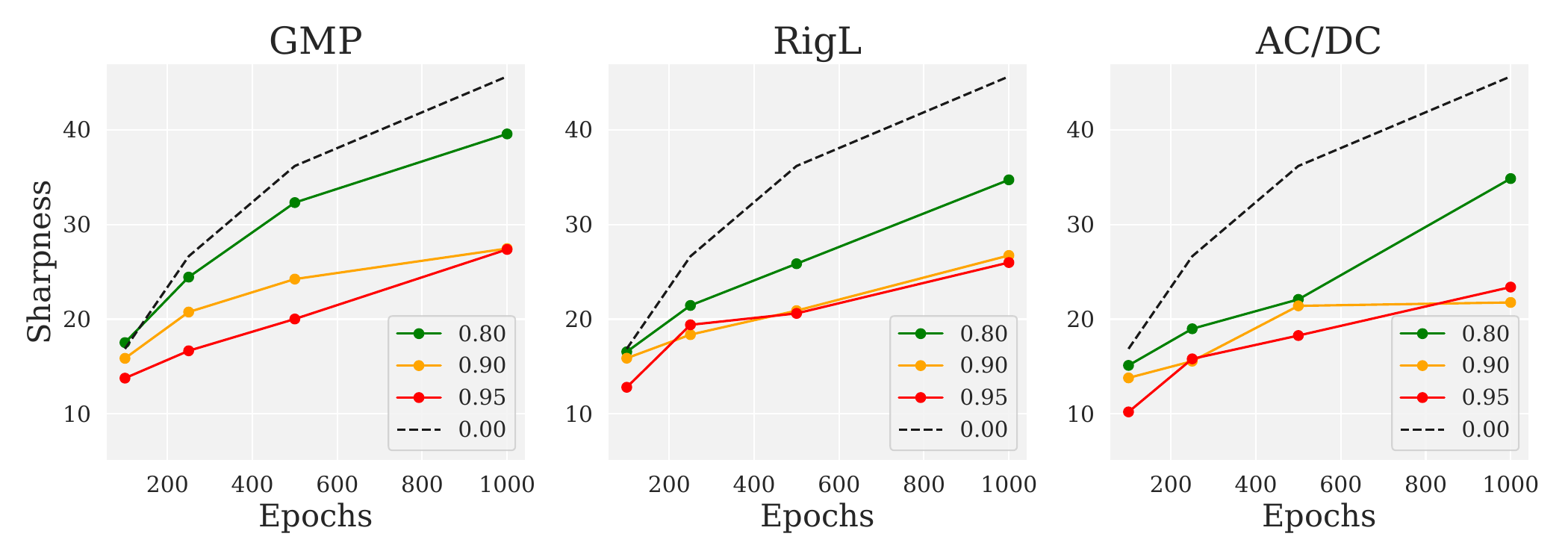}
    \end{subfigure}
    \caption{
       Sharpness (highest eigenvalue) of the loss surface vs number of epochs. 
       Dashed lines correspond to the dense model. 
    }
    \label{fig:asp_sharpness}
\end{figure}

The largest Hessian eigenvalue was estimated via the power iteration method based on Hessian-vector products 
using a customized version of the Eigenthings library~\cite{hessian-eigenthings}. More details about our experimental setup are provided in Appendix 
\ref{appendix:loss_landscape_setup}. 
In Figure~\ref{fig:asp_sharpness}, we observe that, across all methods, sharpness increases with the length of the training run, indicating that sharper minima require extended training to be reached via SGD. Additionally, sharpness decreases with the increase of sparsity. All sparse training methods attain lower sharpness compared to the dense model. 
Models trained with AC/DC and RigL have slightly lower sharpness compared to models trained with GMP, presumably because the former two methods manage to reach flatter optima which are conjectured to have better generalization properties~\cite{merity2017regularizing}.

\begin{figure}[!h]
    \centering
       \begin{subfigure}{0.49\linewidth}
        \includegraphics[width=\linewidth]{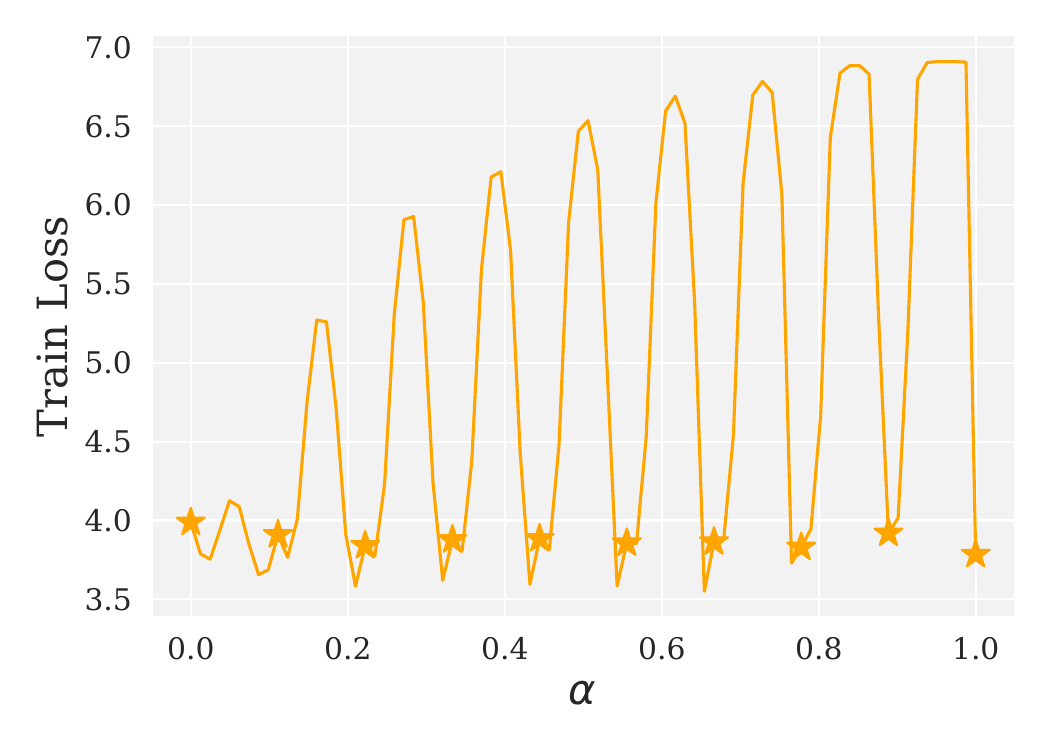}
    \end{subfigure}
    \centering
       \begin{subfigure}{0.49\linewidth}
        \includegraphics[width=\linewidth]{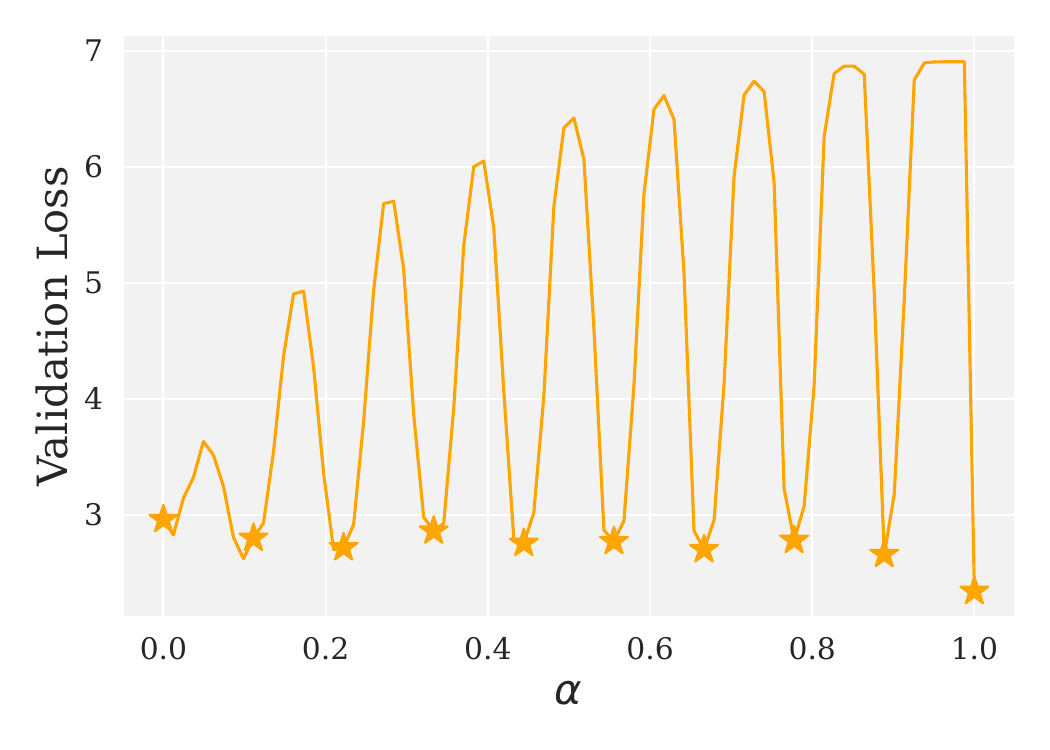}
    \end{subfigure}
    \caption{
        Loss interpolation curves on the training (\textbf{left}) and validation (\textbf{right}) checkpoints of the 95\% sparse 1000-epoch AC/DC model. $\alpha$ corresponds to the fraction of paths traversed from the first checkpoint to the last. Stars denote checkpoints between which loss is interpolated.
    }
    \label{fig:asp_interpolation}
\end{figure}

To examine mode connectivity behavior, in Figure~\ref{fig:asp_interpolation}, we connected the checkpoint obtained on the $99^{th}, 199^{th}, \ldots 999^{th}$ epoch via piecewise-linear curves.  A notable observation is that all checkpoints are separated by a loss barrier, whose height is increasing 
with the number of epochs. Probably, this behavior is a manifestation of the progressive
sharpening phenomenon \cite{cohen2022gradient} where the model sharpness
is increasing gradually with training until reaching the peak value and then plateaus.
Yet, for sparse models, the duration of the longest runs is not long enough to reach the sharpness plateau.

\subsection{Additional quality evaluation of AC/DC++}
\label{sec:additional_quality}

Having demonstrated that extended training has a strong positive effect on sparse model top-1 test accuracy, we further investigate the impact of extended training on other aspects of model quality. We consider two additional quality metrics: their performance in transfer learning scenarios and robustness to common image perturbations.

\cite{iofinova2022transfer} demonstrated that equally sparse models with comparable performance on the original task can vary widely in their performance once finetuned on other, smaller transfer tasks. We compare the transfer performance of dense and 95\% sparse AC/DC++ models, both trained for 100 and 1000 epochs in two transfer learning regimes: linear finetuning, where the hidden layers of the model are trained only on the larger (ImageNet) task, and only the final FC layer is trained on the transfer task, and full-network finetuning, where all layers are finetuned on the transfer task. We find that extended training improves the transfer performance for both transfer scenarios for 95\% sparse models, but is largely neutral for dense models. Full details of the experiment and evaluation are given in Appendix~\ref{appendix:vision_transfer}.

We test robustness by measuring model performance on the ImageNet-C dataset~\cite{hendrycks2019robustness}, which digitally adds 19 types of perturbations to the ImageNet-1K validation set. \cite{liebenwein_lost_2021} and \cite{hooker2019compressed} have found that compressed models are less robust under many types of perturbations, compared to dense models. As before, we consider dense and 95\% sparse AC/DC++ models trained for 100--1000 total epochs. We find that robustness to perturbations increases with training time for sparse models, but stays the same for dense ones. Full details of the experiment and evaluation are given in Appendix~\ref{appendix:robustness_evaluation}.

\section{The Difficulty of Sparse Transfer in Language Modelling}

Motivated by our findings for computer-vision models in previous sections, we extend the analysis to language models, specifically to the very common scenario in which a large language model (BERT-base) is adapted to a specific task via finetuning. In contrast to Section~\ref{sec:additional_quality}, where we examined the effect of extended training on the quality of features created in upstream training, here we examine the impact of sparsity on the optimal recipe for the task of finetuning the model on the downstream dataset.

This setup naturally leads to the following questions: \emph{``do finetuned sparse language models suffer from being undertrained on the downstream task?''}, and \emph{``if yes, does the simple recipe of extended training suffice to mitigate the issue?''}. In this section, we will show that when dense finetuning recipes are used for sparse transfer learning in language models, the resulting models are indeed undertrained and have poor transfer performance. However, we also note an additional difficulty: extended training does not suffice to mitigate the issue, because sparse language models quickly shift from being undertrained to an overfitting regime. The latter is a far larger problem in language understanding tasks than in visual ones, which is likely why we don't observe the same issues with visual transfer learning in Appendix~\ref{appendix:vision_transfer} - there we simply use a long finetuning schedule in all cases. In this section, we explore the problem of balancing under- and over-training in sparse language models and propose a sparse finetuning recipe for creating properly tuned sparse models.%

\subsection{Under Standard Dense Transfer Learning Recipes, Sparse Models are Undertrained}

\paragraph{Experimental Setup.} 
In our experiments, we make use of open-sourced \emph{sparse} pre-trained BERT-base models obtained by~\cite{kurtic2022optimal}. On top of these, we apply various transfer learning recipes to obtain fine-tuned sparse models on datasets from the popular GLUE benchmark~\cite{wang2018glue}.  For fair comparisons with results from prior work, we employ early stopping for all methods. We provide more details about each dataset in Appendix~\ref{appendix:glue}.

The most popular and widely adopted dense transfer learning recipe consists of fine-tuning all weights with linearly decaying learning rate for as much as two or three epochs on the target downstream task. In Table~\ref{tab:dense_recipe} we present results obtained with this approach when applied to sparse models, and denote it as a \textit{dense-transfer recipe}. 
Under the same transfer learning recipe, we clearly observe significant gaps (up to 14 accuracy points on RTE and CoLA)  between the transfer accuracy of the \emph{dense model} (\textit{Dense BERT-base}), and the transfer accuracy of the \emph{sparse model} (\textit{Dense-transfer recipe}).   

\subsection{Extended Training Shifts from Undertraining to Overfitting}
\label{sec:extended}
Observing that the dense transfer learning recipe does not produce competitive sparse finetuned models, we attempt to scale the length of the recipe to mitigate undertraining.
Surprisingly, for sparse language models, this simple technique does not yield a unique setup with consistently better results as models quickly shift from undertraining to an overfitting regime, in which training loss goes to zero, while validation accuracy decreases sharply. To demonstrate this overfitting effect with the extended recipe, in Table~\ref{tab:dense_recipe}, we compare results obtained with this approach (\textit{Extended dense-transfer recipe}) against results obtained by doing a full sweep of finetuning runs with rescaled recipes to \#epochs $\in \{1, 2, 3, ..., \textrm{extended}-1\}$ (\textit{Full sweep of rescaled recipes}). 

\begin{table}[h]
    \centering
    \caption{Sparse-transfer performance of 90\% sparse pre-trained BERT-base model on the dev-set of the corresponding GLUE task, obtained with dense and extended dense (\#epochs=8) transfer learning recipes, as well as with the full sweep of rescaled recipes (\#epochs $\in \{1, 2, ..., 7\}$).}
    \label{tab:dense_recipe}
    \begin{tabular}{c|cccccccc}
         Sparse-transfer & \makecell{RTE\\Acc} & \makecell{QNLI\\Acc} & \makecell{MRPC\\Acc} & \makecell{SST-2\\Acc} & \makecell{CoLA\\Mcc} & \makecell{STS-B\\Pear} & \makecell{MNLI\\ Acc} & \makecell{QQP\\Acc} \\
        \hline
        Dense BERT-base (baseline) &  66.1 & 91.3 & 85.5 & 93.0 & 56.8 & 88.9 & 84.6 & 91.5 \\
        \hline
        Dense-transfer recipe & 52.4 & 88.9 & 82.8 & 91.2 & 42.5 & 87.1 & \textbf{82.2} & 90.0 \\
        Extended dense-transfer recipe & 55.2 & 88.7 & \textbf{85.6} & 91.4 & 47.2 & 87.6 & 81.6 & 90.3 \\
        Full sweep of rescaled recipes & \textbf{\makecell{57.0}} & \textbf{\makecell{89.3}} & \makecell{84.1} & \textbf{\makecell{92.0}} & \textbf{\makecell{48.5}} & \textbf{\makecell{88.0\\}} & \textbf{\makecell{82.2}} & \textbf{\makecell{90.4}} \\
        \hline
        Best recipe length & 5 ep & 2 ep & 5 ep & 2 ep & 7 ep & 4 ep & 3 ep & 5 ep \\
    \end{tabular}
\end{table}

The results suggest that with the existing recipes, there is no one-size-fits-all solution. Versions of this rescaling approach have been utilized by prior works like \cite{kurtic2022optimal} and \cite{zafrir2021prune} to obtain accurate sparse models on various downstream datasets. However, this approach comes with a huge computational burden: for each rescaled recipe, a full hyperparameter sweep over relevant parameters has to be done in order to obtain competitive finetuned sparse models.
Due to practicality and associated costs, this is not a desirable solution in practice.

\subsection{Sparse Transfer Learning for Language Models}
\label{sec:BERT_transfer}

In the previous section, we have demonstrated the following three problems with the existing approach of either using the dense finetuning recipe, or simply extending it for sparse finetuning:
\begin{enumerate}
    \item following dense-transfer recipes, sparse language models are undertrained;
    \item even at high sparsities, these models can still exhibit overfitting behavior under the extended training regime; 
    \item finding the optimal recipe to mitigate undertraining and overfitting has major computational burdens.
\end{enumerate}

To address these issues, we propose a simple approach for sparse transfer in NLP, which produces highly accurate and competitive sparse models on a wide range of downstream datasets with minimal hyperparameter tuning. Our technique is  inspired by the idea of gradual layer unfreezing presented in the ULMFiT framework~\cite{howard2018universal}, which introduced a universal framework for fine-tuning \emph{dense} language models for text-classification tasks, with a focus on LSTM models~\cite{merity2017regularizing}. Based on ULMFiT and findings of \cite{yosinski2014transferable}, which suggests that different layers capture different information and therefore should be fine-tuned to different extents, we adopt the idea of gradual unfreezing and adjust it for \emph{transformer-based~\cite{vaswani2017attention} sparse} language models.

More specifically, we focus on the popular BERT-base model which consists of three groups of layers: embeddings, 12 identical transformer blocks, and a task-specific classifier head. Sparsified versions of this model, which are the main interest of this work, prune all linear layers across all transformer blocks, which is the standard practice in literature~\cite{2020-sanh, kurtic2022gmp, kurtic2022optimal, zafrir2021prune} and brings the best accuracy-vs-latency trade-offs~\cite{kurtic2022optimal}.

\begin{table}[t]
    \centering
    \caption{Our sparse-transfer performance of 90\% sparse pre-trained BERT-base model on the dev-set of the corresponding GLUE tasks, benchmarked against the current state-of-the-art sparse-transfer results from Prune OFA~\cite{zafrir2021prune} and oBERT~\cite{kurtic2022optimal}.} 
    \label{tab:bert-sparse-transfer}
    \resizebox{\textwidth}{!}{
    \begin{tabular}{c|cccccccc}
        Sparse-transfer & \makecell{RTE\\Acc} & \makecell{QNLI\\Acc} & \makecell{MRPC\\F1 / Acc} & \makecell{SST-2\\Acc} & \makecell{CoLA\\Mcc} & \makecell{STS-B\\Pear / Spear} & \makecell{MNLI\\m / mm} & \makecell{QQP\\Acc / F1} \\
        \hline
        Dense BERT-base &  66.1 & 91.3 & 89.8 / 85.5 & 93.0 & 56.8 & 88.9 / 88.5 & 84.6 / 83.4 & 91.5 / 88.5 \\
        \hline
        Prune OFA~\cite{zafrir2021prune} & N/A & 89.1 & N/A & 90.9 & N/A & N/A & 81.5 / 82.4 & \textbf{90.9} / \textbf{87.6} \\ 
        oBERT~\cite{kurtic2022optimal} & 57.0 & 89.3 & 89.3 / \textbf{85.6} & \textbf{92.0} & 48.5 & \textbf{88.0 / 87.6} & 82.2 / 82.5 & 90.4 / 87.1 \\
        This work & \textbf{60.1} & \textbf{90.5} & \textbf{89.7} / 85.2 & 91.8 & \textbf{51.4} & 87.2 / 87.1 & \textbf{83.7} / \textbf{83.8} & \textbf{90.9} / \textbf{87.6} \\
    \end{tabular}
    }
\end{table}

\begin{figure}[!h]
    \centering
    \begin{subfigure}{0.49\linewidth}
        \includegraphics[width=\linewidth]{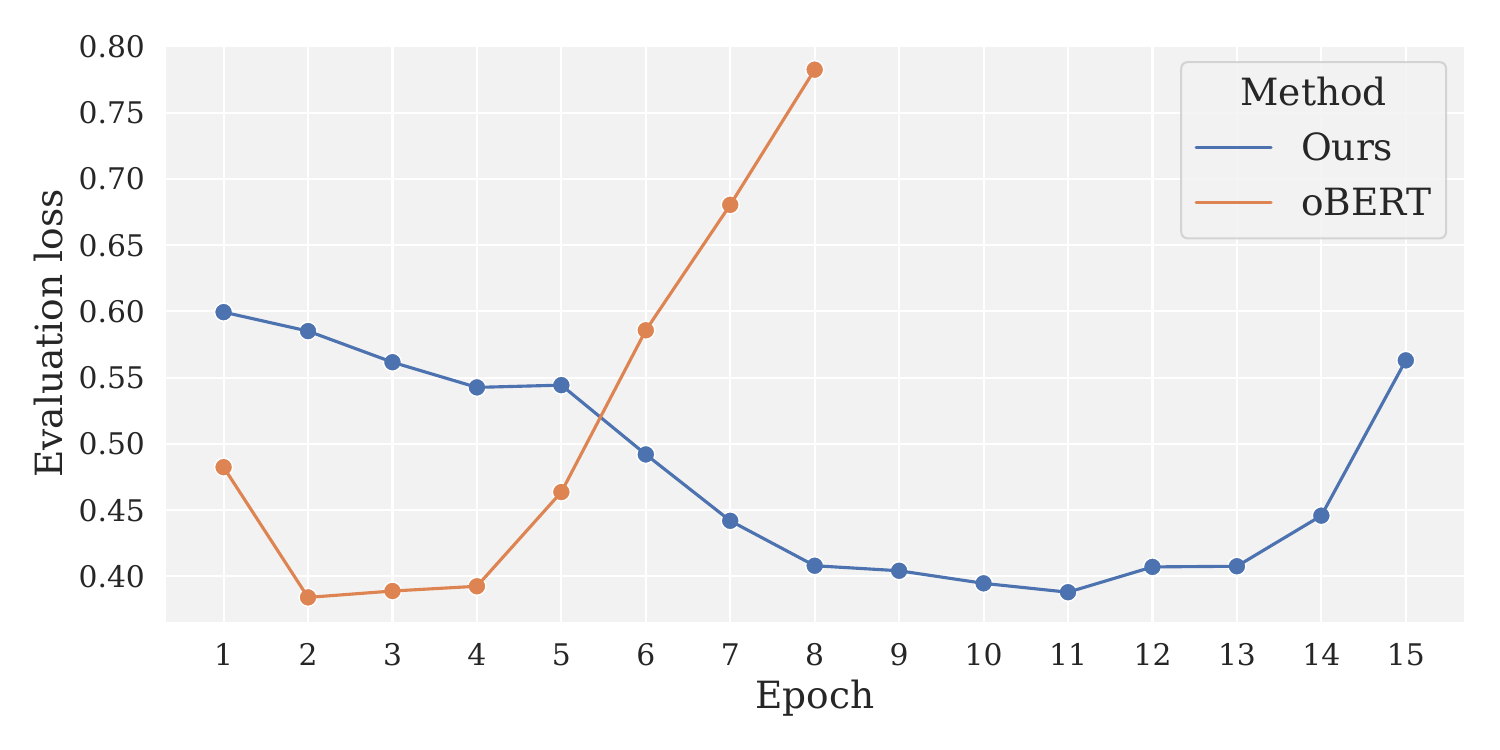}%
    \end{subfigure}
       \begin{subfigure}{0.49\linewidth}
        \includegraphics[width=\linewidth]{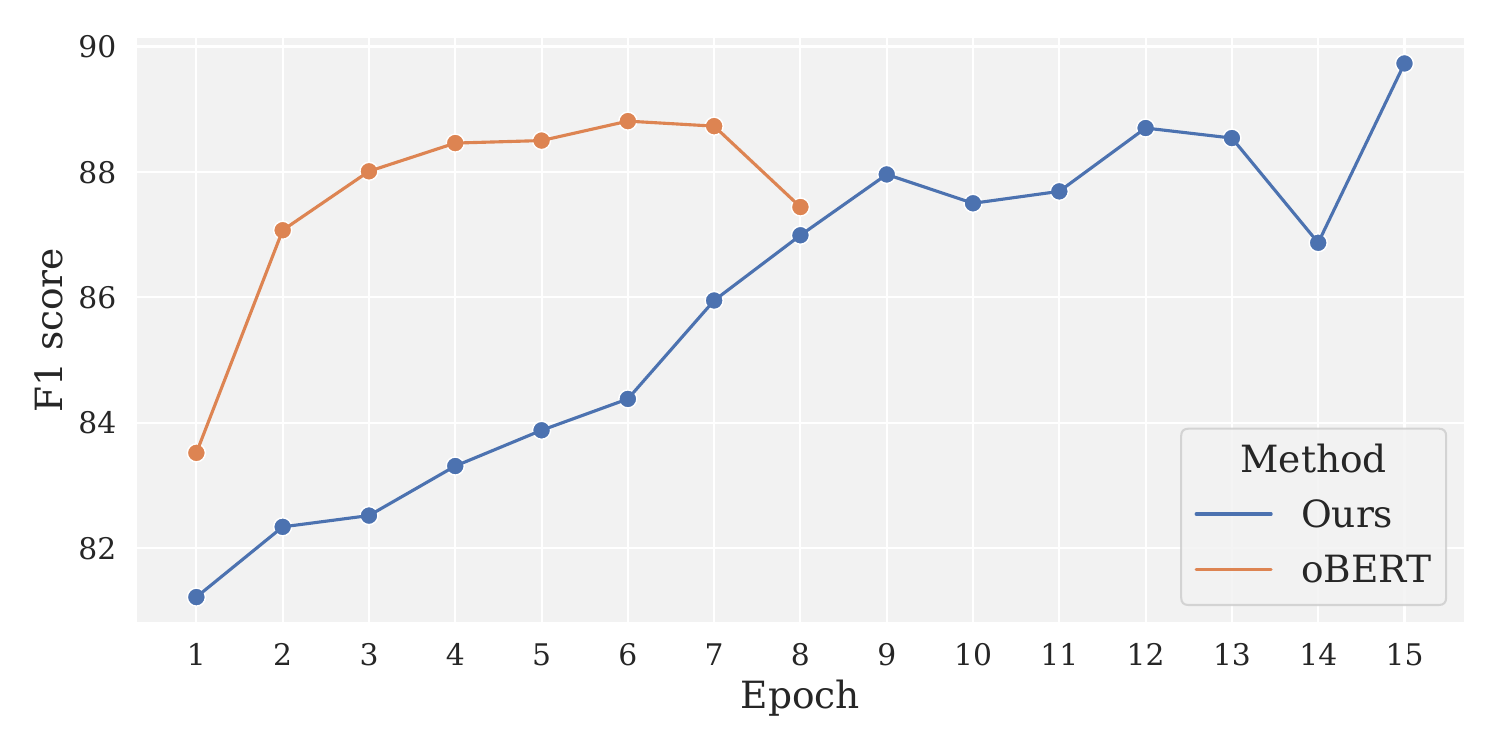}%
    \end{subfigure}
    \caption{
       Evaluation loss (lower is better) and F1 score (higher is better) during sparse-transfer with oBERT~\cite{kurtic2022optimal} and our approach on MRPC dataset.
    }
    \label{fig:mrpc}
\end{figure}

\begin{figure}[!h]
    \centering
    \begin{subfigure}{0.49\linewidth}
        \includegraphics[width=\linewidth]{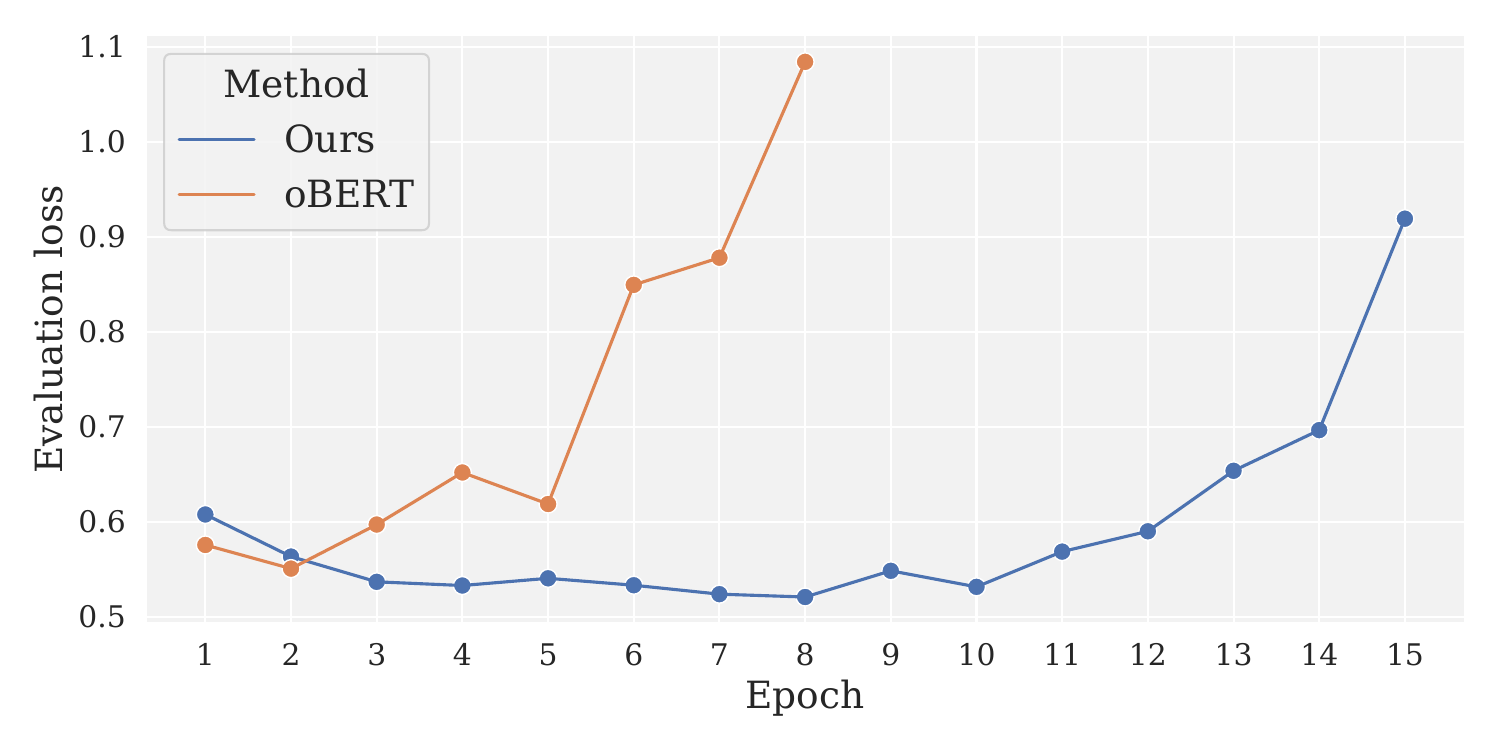}%
    \end{subfigure}
       \begin{subfigure}{0.49\linewidth}
        \includegraphics[width=\linewidth]{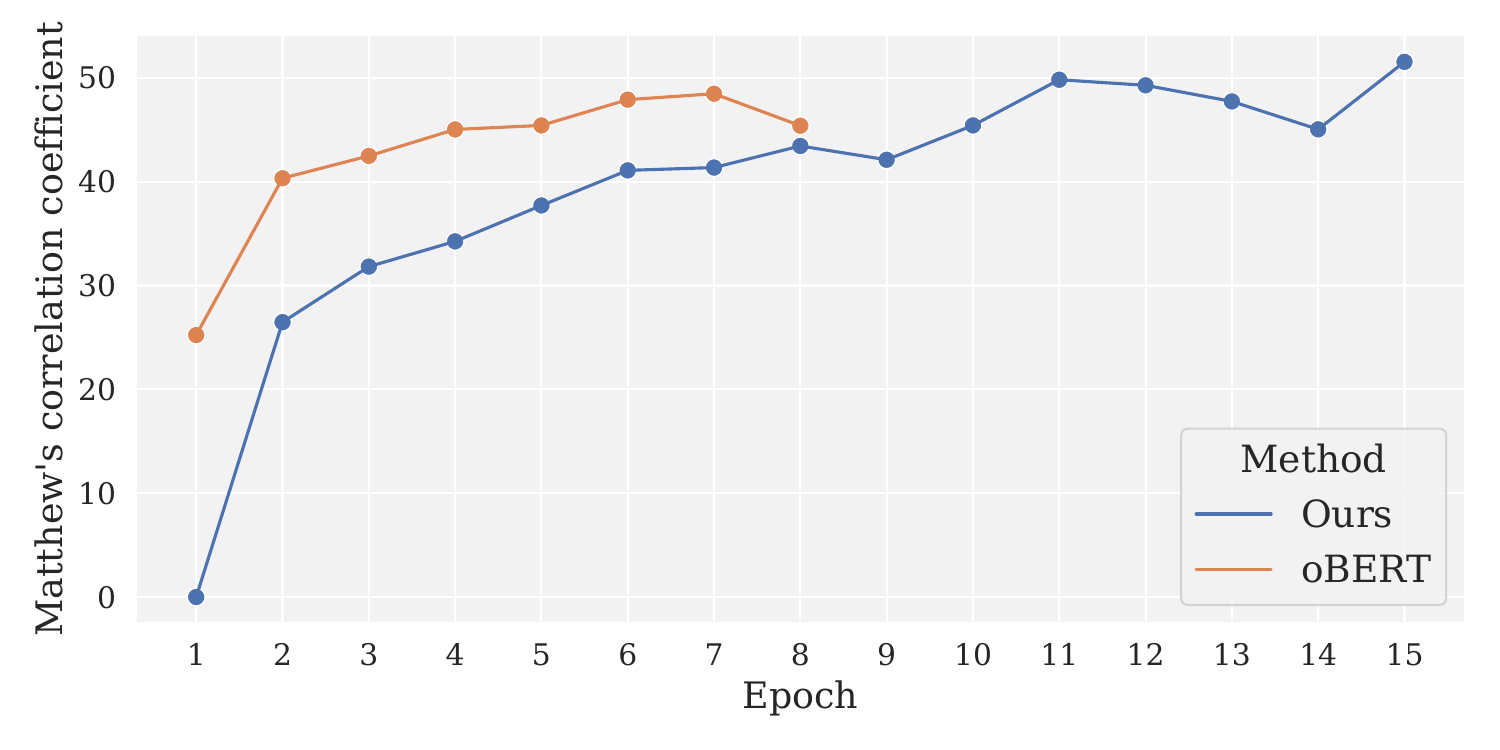}%
    \end{subfigure}
    \caption{
       Evaluation loss (lower is better) and Matthew's correlation coefficient (higher is better) during sparse-transfer with oBERT~\cite{kurtic2022optimal} and our approach on CoLA dataset.
    }
    \label{fig:cola}
\end{figure}

Our approach can be summarized as follows. For each downstream task, we start from a sparse pre-trained model produced by~\cite{kurtic2022optimal} and randomly initialize a task-specific classifier head. Then we freeze all embeddings and sparsified linear weights, while keeping their biases and corresponding LayerNorm~\cite{ba2016layer} layers unfrozen and trainable. We start by finetuning only the classifier head and all other trainable parameters (biases and LayerNorms) for one epoch, and then follow the same process from back-to-front by unfreezing the unpruned linear weights in preceding transformer blocks. After the last layer is unfrozen and finetuned, we continue finetuning all layers together for one more epoch.

Given that at each epoch we have a different model architecture (one more sparse transformer block unfrozen relative to the previous epoch), we finetune it with the linearly decaying learning rate and then rewind back to the initial value for the next epoch. We have also tried the slanted triangular learning rate schedule proposed in ULMFiT, but we found the warmup phase not very helpful as it is known that sparse language models usually require much higher learning rates relative to their dense counterparts in order to train and converge successfully~\cite{kurtic2022gmp}.

To validate the effectiveness of our proposed sparse transfer approach, we benchmark it against the two current state-of-the-art sparse-transfer results presented in \emph{Prune Once for All (Prune OFA)}~\cite{zafrir2021prune} and \emph{The Optimal BERT Surgeon (oBERT)}~\cite{kurtic2022optimal} papers. The former makes use of knowledge distillation from a finetuned dense teacher model, while the latter uses a full sweep over extended and rescaled dense transfer recipes, such as the ones we presented in Section~\ref{sec:extended}. As can be seen from Table~\ref{tab:bert-sparse-transfer}, our approach outperforms highly competitive results by Prune OFA in all, and oBERT in eight out of twelve datasets, setting new state-of-the-art accuracy-vs-sparsity results for many tasks in the GLUE benchmark suite. It is worth emphasizing that all of our results are obtained with significantly less hyperparameter tuning than the other two competing methods, which aligns with our goal of finding a stable one-size-fits-all solution for the sparse-transfer problem. We search and tune the initial learning rate in \{1e-4, 2e-4, 3e-4\}, and dropout in \{0.05, 0.1\}, and report mean performance over the two best runs. Thus, our grid consists of only 6 different combinations for each considered dataset, whereas competing approaches sweep over 54 (~\cite{zafrir2021prune}) and 24 (~\cite{kurtic2022optimal}) different combinations. It is worth emphasizing that all of the considered methods, including ours, have noticeable variability in results on small datasets across different seeds and hyperparameter configurations, which aligns with findings of ~\cite{devlin2018bert}.

To better understand what happens during our proposed sparse transfer learning setup, and to develop an intuition about why it is able to provide stable and competitive results across many different datasets ranging in sizes from 2.4k (RTE) and 392k (MNLI) labeled samples, we visualize evaluation loss and evaluation accuracy metrics over the entire transfer learning process in Figures~\ref{fig:mrpc} and~\ref{fig:cola}. As can be seen, our approach enables slower and therefore more stable transfer learning on the target datasets which effectively prevents overfitting, even though the total number of epochs is two times larger than the extended dense-transfer recipes analyzed in Section~\ref{sec:extended}. This aligns with findings in ULMFiT, which demonstrates that gradual unfreezing in combination with a carefully designed learning rate schedule prevents catastrophic forgetting and enables robust transfer learning across a wide range of different downstream tasks.

\section{Conclusion}

In this work, we examined the impact of high sparsity on model training under standard computer vision and natural language recognition scenarios, and provided evidence that traditional training recipes used for dense models are generally too short for sparse training. Starting from this observation, we were able to produce state-of-the-art models for sparse computer vision on two classic benchmarks for pruning: the ResNet50/ImageNet from-scratch training benchmark, and transfer learning from BERT-base on several NLP datasets. Our work focused on 
the differences between sparse and dense training dynamics and their effect on optimal training, providing additional analysis towards the difficulty of sparse training. The main motivation for our work is to inspire further research in adapting training schedules, hyperparameters, and optimizers to improve sparse model training in order to reach higher accuracies under sparsity, but also to do so efficiently.
We leave this as a challenge to the community.  

\section*{Acknowledgements}

We gratefully acknowledge funding from the European Research Council (ERC) under the European Union’s Horizon 2020 programme (grant agreement No 805223 ScaleML). E.I. was supported in part by the FWF DK VGSCO, grant agreement number W1260-N35.
D.K. was supported by Russian Science Foundation, grant 21-11-00373.

\newpage

\bibliography{bibliography.bib}
\newpage

\appendix

\addcontentsline{toc}{section}{Appendix} %
\part{Appendix} %
\parttoc %

\setcounter{table}{0}
\renewcommand{\thetable}{\Alph{section}.\arabic{table}}
\setcounter{figure}{0}
\renewcommand{\thefigure}{\Alph{section}.\arabic{figure}}

\section{Training hyperparameters}
\label{appendix:hyperparameters}

We adopt and scale the training recipes proposed in the FFCV library \cite{leclerc2022ffcv}. 
All runs use a linear decaying learning rate with a peak value $\eta_{\text{max}} = 0.5$ at 2nd 
epoch that is gradually decreased to 0 at the end of the training. We train the 
models with batch size of 1024. 

\begin{table}[!h]
  \caption{Augmentation and regularization procedure used in the work.}
  \label{tab:training_procedure}
  \centering
  \begin{tabular}{c|c}
  \toprule
  Method & Value \\
  \midrule
  Weight decay & 1e-4 \\
  Label smoothing $\varepsilon$ & 0.1 \\
  Dropout & \XSolidBrush \\
  \midrule
  H.flip & \Checkmark \\
  RRC & \Checkmark \\
  Blurpool & \Checkmark \\
  Progressive resizing  & \Checkmark \\
  \midrule
  Test crop ratio & 0.875 \\
  \bottomrule
  \end{tabular}
\end{table}

We use progressive resizing with training on random resized crops (RRC) of size $160 \times 160$ 
for the first 80\% of training and the remaining 20\% of training with crops of size $192 \times 192$. 
We run classifier on $224 \times 224$ center crops 
for the image resized to $256$ px on validation following the standard ImageNet evaluation protocol.
Differently from the FFCV recipes, we do not use test-time augmentation with averaging 
of the prediction on the original image and its flipped version since it requires 2x more
inference FLOPs. For MobileNet experiments we adopted smaller value of weight decay 3e-5 following 
the \cite{kusupati2020soft} work and learning rate $\eta=1.024$. 

\section{Sparse training algorithm hyperparameters}

All sparse training methods use the same optimizer and learning rate scheduler hyperparameters and differ only 
in the specifics of weight pruning / regrowth procedure. In our work we prune only weights of convolutional
(we do not prune biases and batch norm parameters) and keep the first convolution as well as the classification head 
(the last linear layer outputting the logits for classes) dense. This setup is common in the literature.

\paragraph{AC/DC}

In our AC/DC training setup we train the model without sparsity for the first 10\% of the training and then alternate 
between sparse and dense training each 5 steps following the original paper \cite{peste2021ac}. 
We run the last compression step and last decompression step for 10 and 15 epochs respectively, again following the prescription 
from the paper. We have ablated the duration  of the compression and decompression phases and 5 epochs appeared to yield the best performance.

\paragraph{RigL}

In the original paper \cite{evci2020rigging} authors train the models with fixed target sparsity and periodically 
update the fraction of connections in each pruned layer $l$ following the cosine decay rule:
\begin{equation}
\frac{\alpha}{2} \left(1 + \cos\left(\frac{\pi t}{T}\right)\right) (1 - s_l)
\end{equation}
Above $s_l$ is the sparsity of a given layer $l$ and $\alpha$ is the initial fraction of 
updated connections. The choice of update frequency $\Delta T$ has a strong impact on the method performance 
and in the original work authors obtained the best performance with $\Delta T \simeq 100-300$ steps, which corresponds roughly 
to 0.4-1.2 epochs on ImageNet with the batch size used in the work. In our work we updated connections every epoch
to have setup close to the original work. We set $\alpha=0.3$ as in the one used in the RigL work. 
Following the original work, we train with a fixed sparsity mask for the 
last 25\% of training. For global sparsity we apply the ERK (Erdős–Rényi-Kernel) sparsity profile as in the original paper 
as it was shown to produce the best performance for a given sparsity. 

\paragraph{GMP}

We gradually increase the sparsity from 0\% (dense model) to the target sparsity following a cubic interpolation law. 
Sparsity is increased every 5 epochs. We train with a fixed mask for the last quarter of training duration as in RigL. 

\section{AC/DC dense fraction ablation}

The original AC/DC paper has equal length of compression and decompression phases and one 
may raise a question, whether it would be better to train the model longer in compressed state or decompressed
to achieve better performance for a given sparsity target. 
In the experiments below we train the ResNet-50 model for 1000 epochs with 95\% sparsity.
It turns out that allocating equal time for dense and sparse training works the best, as 
one can see from the Table \ref{tab:sparse_training_fraction}.  
Here 0 means pruning at some moment of training and finetuning with fixed mask 
and 1 is the accuracy of dense model. Even when training most of the time in sparse regime
one still doesn't lose much in performance. 

\begin{table}[!htb]
    \caption{Ablation on the sparse training fraction.}
    \label{tab:sparse_training_fraction}
    \centering
    \begin{tabular}{cc}
        \toprule
        Sparse training fraction (\%) & Accuracy (\%) \\
        \midrule
        0 & 79.0 \\
        0.2 & 77.2 \\
        0.5 & 77.5 \\
        0.8 & 77.3 \\
        1 & 75.0 \\
        \bottomrule
    \end{tabular}
\end{table}

\section{AC/DC phase duration ablation}

Another hyperparameter that may have a significant impact on performance is the update interval of 
AC/DC iteration. Intuitively, more frequent updates may allow for more changes in the choice of sparsity 
of the pruning mask and at the same longer training with fixed mask can be needed to the weights to adapt well 
for a given sparsity pattern. We have tried different choices of AC/DC iteration duration for ResNet-50 trained for 
1000 epochs and 95\% global sparsity and the chosen frequency update of 5 epochs is close to the optimum.

\begin{table}[!htb]
    \caption{AC/DC iteration length ablation.}
    \label{tab:ac_dc_duration}
    \centering
    \begin{tabular}{cc}
        \toprule
        Sparse training fraction (\%) & Accuracy (\%) \\
        \midrule
        1 & 77.0 \\
        5 & 77.5 \\
        10 & 77.2 \\
        50 & 75.9 \\
        \bottomrule
    \end{tabular}
\end{table}

\section{Comparison with the small dense model}

One may ask whether one can achieve similar performance 
with a smaller dense model. We have taken a smaller version of ResNet50 with 2x smaller 
hidden dimension (denoted ResNet-50x0.5). We trained the model with the same training procedure
as the baseline ResNet-50 model for 1000 epochs. One can observe from Table~\ref{tab:small_dense_vs_sparse} that even for smaller number of 
FLOPs sparse model is much better than the dense one. 

\begin{table}[!htb]
    \caption{Small-dense vs sparse model on ImageNet-1k}
    \label{tab:small_dense_vs_sparse}
    \centering
    \begin{tabular}{cccc}
        \toprule
        Model (\%) & Sparsity (\%) & FLOPs & Accuracy (\%) \\
        \midrule
        ResNet-50x0.5 & 0 & - & 69.8 \\
        ResNet-50 & 95 & - & 77.5 \\
        \bottomrule
    \end{tabular}
\end{table}

\section{Impact of extended training on transfer performance}
\label{appendix:vision_transfer}

We further validate our results by using sparse and dense ResNet50 models pretrained on ImageNet for transfer learning for other vision recognition tasks. Transfer learning is a common paradigm in which a large dataset is used to set network weights before they are further fine-tuned on the (typically, smaller) dataset of interest; this approach can provide significant gains over direct training on the smaller task from a random initialization. In this section, we refer to the larger dataset/task (in our case, ImageNet) as the \emph{upstream} task, and the smaller dataset/task as the \emph{downstream} task.

There are two common approaches to transfer learning, the choice of which largely depends on the technological capabilities of the system doing the learning task. If a large amount of compute is available, all model weights can be trained on the downstream task, in a process we call \emph{full finetuning}. Otherwise, if compute is limited, all the layers of the deep neural network except for the final classifier are frozen after downstream training and used purely as a feature extractor, and only the final layer (properly resized) is trained on the downstream task, as a linear classifier on the extracted features. We call this process \emph{linear finetuning}.

We apply both transfer learning approaches to a standard set of twelve downstream tasks, which are frequently used as benchmarks for transfer learning performance\cite{kornblith2019better, iofinova2022transfer, salman2020adversarially}. The twelve datasets include six specialized tasks and six general tasks; a full list of downstream tasks is given in Appendix Table~\ref{table:datasets}.

We use four variants of ResNet50 models pretrained on ImageNet as the upstream model: dense and 95\% globally sparse models trained for 100 and 1000 epochs. Otherwise, we follow the training hyperparameters of \cite{iofinova2022transfer}, and, also following this work, we compute a single metric across all twelve tasks by computing Average Increase in Error for a set of tasks $T$ over a baseline model. This metric is computed as follows: for each of the downstream tasks, we compare the difference in the error ($1 - $Top-1 accuracy) when the baseline (100-epoch dense) model is used for transfer learning, versus when another model (either sparse, or extended-training dense) is used. The increase in error is the difference of these two quantities divided by the dense error; the final metric is the average of these relative differences across all downstream datasets. As argued in \cite{iofinova2022transfer}, normalizing error differences by the dense error allows us to roughly equalize dataset impact when we compute a single metric across different datasets with very different model performance (e.g., nearly 100\% accuracy on the Pets dataset, but about 50\% accuracy on the Aircraft dataset. Formally, the metric is computed as:
\begin{equation}
    \textit{AIE} = \frac{1}{|T|}\sum_{t \in T}\frac{\textit{Err}_{\textit{Model}, t} - \textit{Err}_{\textit{Baseline}, t}}{\textit{Err}_{\textit{Baseline}, t}}
\end{equation}

We compute the AIE using the 100-epoch dense model as the baseline model, and present the results in Table~\ref{tab:imagenet_transfer}. We observe that the effect of extended training on the dense model is neutral at best - there is a small increase in error with extended training in the linear finetuning regime, and no change in the full finetuning regime. Conversely, the 1000 epoch 95\% sparse model outperforms the 100-epoch one across both regimes, with a higher error drop in the linear finetuning regime and a smaller error increase in the full finetuning regime.

\begin{table}[!htb]
    \centering
    \begin{tabular}{c|cc|cc}
        \toprule
        Training Duration & Linear - Dense & Linear - 95\% Sparse & Full - Dense & Full - 95\% Sparse \\
        \midrule
        100ep. & - & -0.123 & - & 0.112\\
        1000ep. & 0.026 & -0.182 & 0.000 & 0.062\\
        \bottomrule
    \end{tabular}
    \caption{Transfer learning for image recognition - Average Increase in Error over the 100-epoch dense model. Lower is better: positive numbers indicate that the model has, on average, more error than the baseline model; negative numbers indicate that the model has, on average, less error.}
     \label{tab:imagenet_transfer}
\end{table}

\begin{table}[h]
\centering
\scalebox{0.8}{
\begin{tabular}{@{}ccccc@{}}
\toprule
Dataset & Number of Classes &  Train/Test Examples & Accuracy Metric\\
\midrule
SUN397\cite{xiao2010SUN} & 397 & 19 850 / 19 850 & Top-1 \\
FGVC Aircraft\cite{maji13fgvc-aircraft} & 100 & 6 667 / 3 333 & Mean Per-Class \\
Birdsnap\cite{berg2014birds} & 500 & 32 677 / 8 171 & Top-1 \\
Caltech-101\cite{li2004caltech101} & 101 & 3 030 / 5 647 & Mean Per-Class \\
Caltech-256\cite{holub2006caltech256} & 257 & 15 420 / 15 187 & Mean Per-Class \\
Stanford Cars\cite{krause2013cars} & 196 & 8 144 / 8 041 & Top-1 \\
CIFAR-10\cite{cifar100} & 10 & 50 000 / 10 000 & Top-1 \\
CIFAR-100\cite{cifar100}  & 100 & 50 000 / 10 000 & Top-1 \\
Describable Textures (DTD)\cite{cimpoi2014dtd} & 47 & 3 760 / 1 880 & Top-1 \\
Oxford 102 Flowers\cite{nilsback2006flowers} & 102 & 2 040 / 6 149 & Mean Per-Class \\
Food-101\cite{bossard2014food101} & 101 & 75 750 / 25 250 & Top-1 \\
Oxford-IIIT Pets\cite{parkhi2012apets} & 37 & 3 680 / 3 669 & Mean Per-Class \\
\bottomrule
\end{tabular}
}
\caption{Datasets used as downstream tasks for transfer learning with computer- vision models. }
\label{table:datasets}
\end{table}

\begin{figure}[h]
    \centering
    \begin{subfigure}{0.49\linewidth}
        \includegraphics[width=\linewidth]{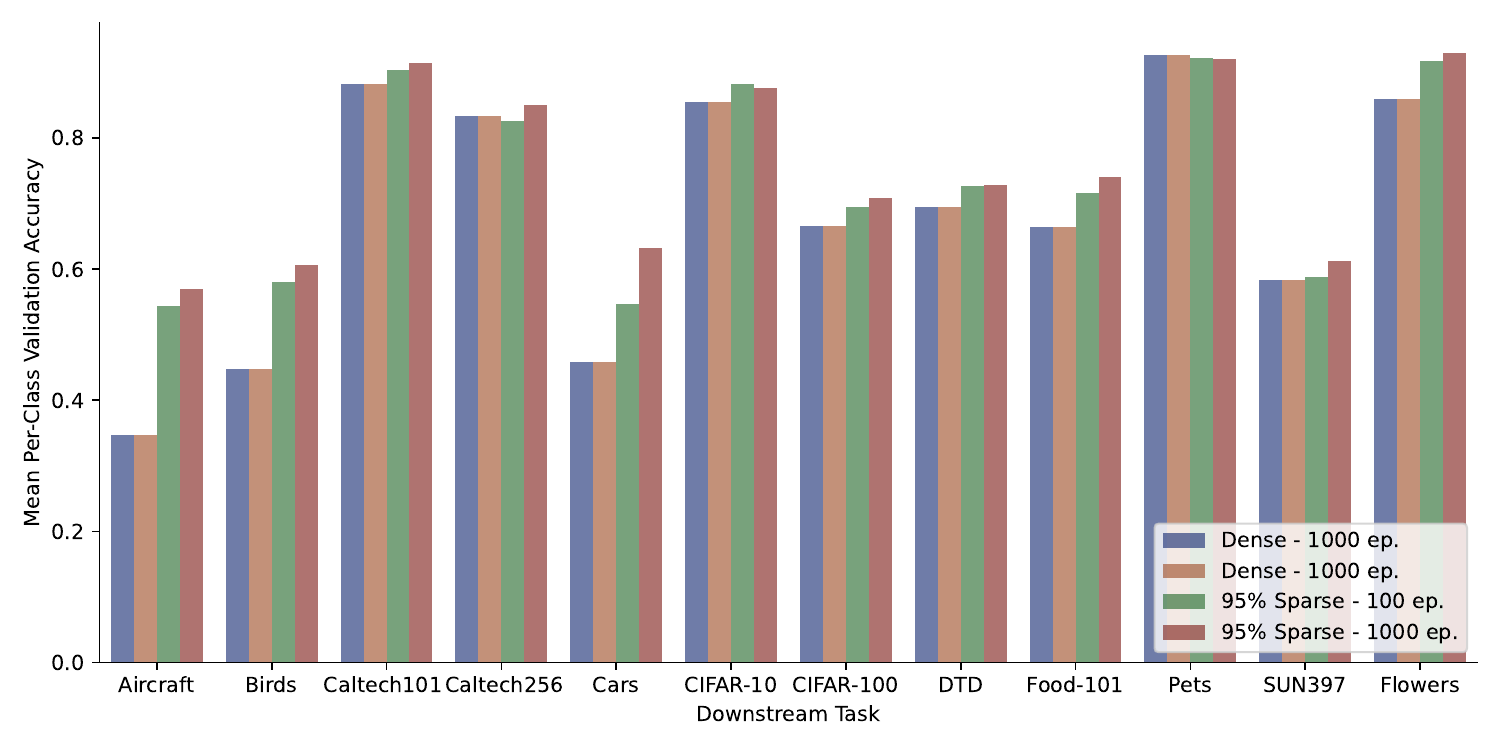}
    \end{subfigure}
       \begin{subfigure}{0.49\linewidth}
        \includegraphics[width=\linewidth]{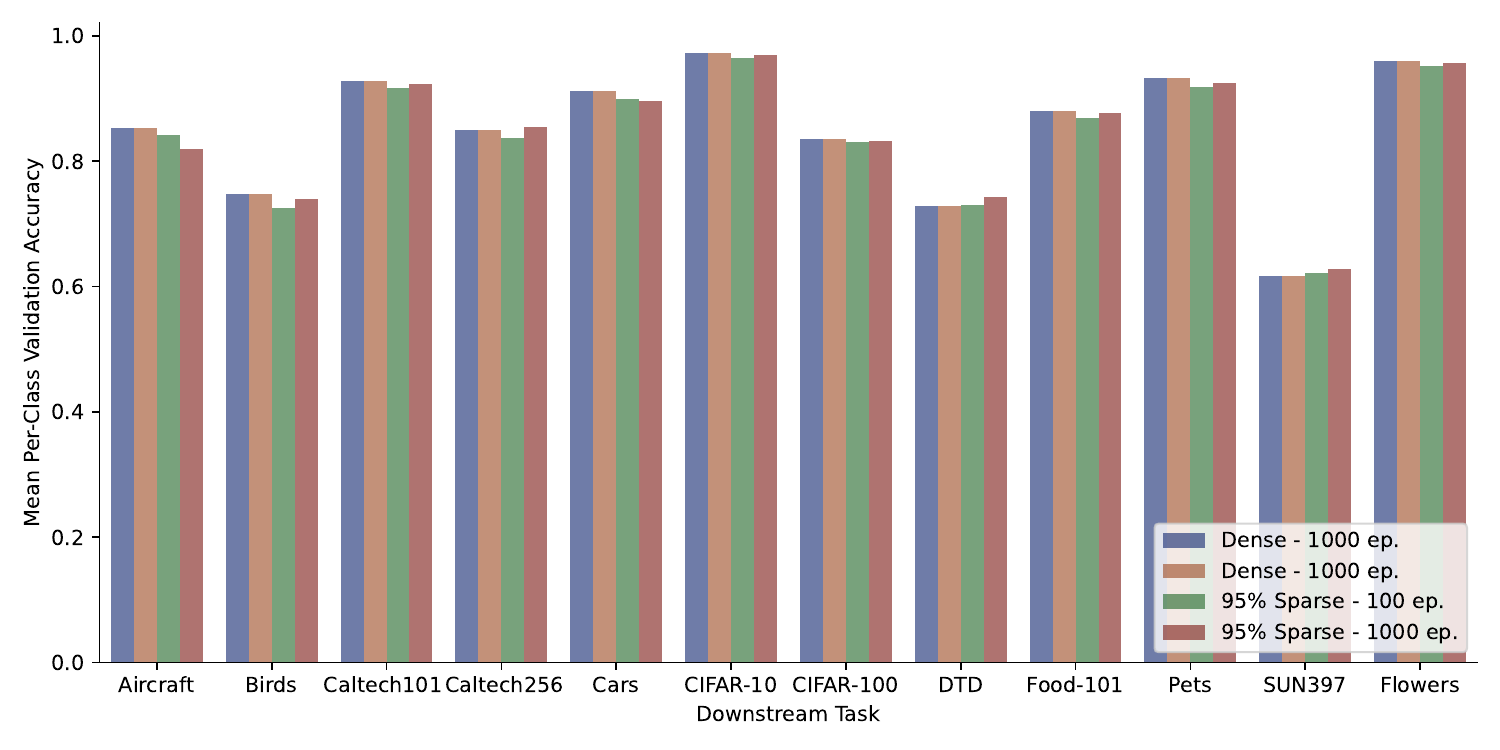}
    \end{subfigure}    
    \caption{Linear (left) and Full-network (right) transfer learning mean per-class accuracy for individual datasets.
    }
    \label{fig:transfer_comparison}
\end{figure}

\section{Evaluation on ImageNet-C}
\label{appendix:robustness_evaluation}
\begin{figure}[h]
    \centering
    \begin{subfigure}{0.85\linewidth}
        \includegraphics[width=\linewidth]{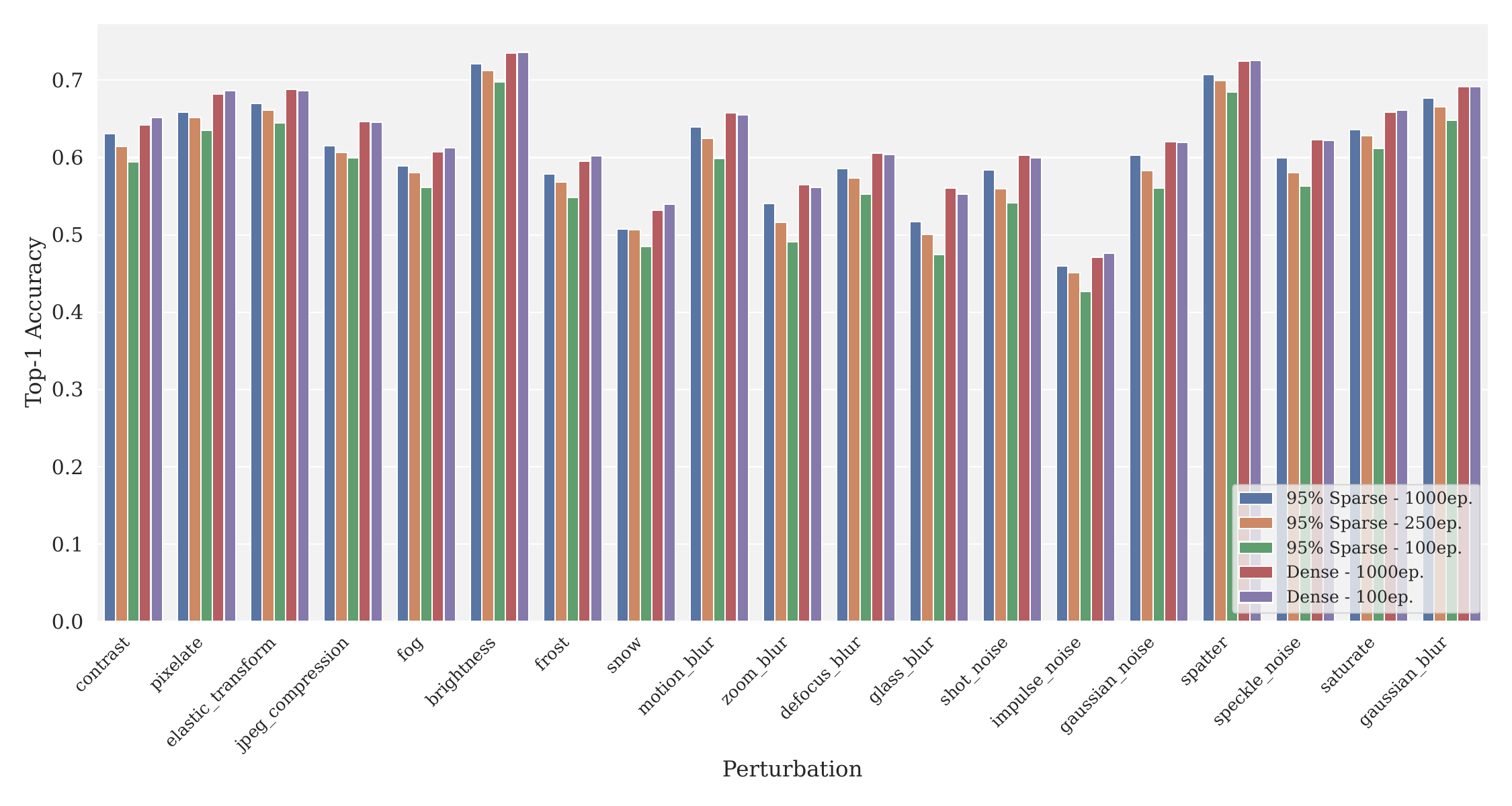}
    \end{subfigure} 
    \caption{Linear (left) and Full-network (right) transfer learning mean per-class accuracy for individual datasets.
    }
    \label{fig:imagenet_c_1}
\end{figure}

In order to confirm the robustness of our models against perturbations, we evaluate AC/DC and dense models on the ImageNet-C\cite{hendrycks2019robustness} dataset, which consists of the standard ImageNet validation dataset to which perturbations such as noise, blur, photographic effects (such as contrast enhancement) or weather conditions (such as snow or fog) were digitally added. We use the lowest ``1'' level of perturbation, as we find that the quality already drops considerably over the clean data. Our results are shown in Figure~\ref{fig:imagenet_c_1}. We observe that, as expected and also as reported in \cite{liebenwein_lost_2021}, dense models outperform sparse ones on corrupted data. Further, and consistently with our other findings, extended training on the dense model does not improve performance on the ImageNet-C dataset; conversely, such improvement can be seen for all 19 perturbation categories for 95\% sparse models.

\section{Transfer learning datasets for language models}
\label{appendix:glue}

In Table~\ref{tab:glue} we provide a brief summary of datasets in the popular GLUE benchmark~\cite{wang2018glue}. Following previous work~\cite{sanh2020movement, zafrir2021prune, kurtic2022gmp, kurtic2022optimal, zhang2022platon, huang2021sparse}, we exclude WNLI dataset from consideration in all experiments.

\begin{table}
\centering
\begin{tabular}{@{}ccccc@{}}
\toprule
Dataset & Train/Test Examples & Accuracy Metric\\
\midrule
RTE~\cite{wang2018glue} & 2.5k / 3k & Accuracy \\
QNLI~\cite{wang2018glue} & 105k / 5.4k & Accuracy \\
MRPC~\cite{mrpc} & 3.7k / 1.7k & F1 score and Accuracy \\
SST-2~\cite{sst2} & 67k / 1.8k & Accuracy \\
CoLA~\cite{cola} & 8.5k / 1k & Matthews correlation coefficient \\
STS-B~\cite{stsb}& 7k / 1.4k & Pearson and Spearman correlation \\
MNLI~\cite{mnli} & 393k / 20k & Matched (m) and mismatched (mm) accuracy \\
QQP~\cite{wang2018glue} & 364k / 391k & Accuracy and F1 score \\
\bottomrule
\end{tabular}
\caption{Datasets used as downstream tasks for transfer learning with language models. }
\label{tab:glue}
\end{table}

\section{Loss landscape analysis experimental setup}
\label{appendix:loss_landscape_setup}

To compute the largest Hessian eigenvalue we ran power iteration method \cite{MisesPraktischeVD}
for 20 iterations using the modified version of the code from \cite{hessian-eigenthings} package. 
20 iterations were usually enough for the iterations to converge. 
Hessian is computed only with respect to the non-zero weights.

To plot \ref{fig:asp_interpolation} we evaluated training loss (with augmentation turned off)
and validation loss on the whole dataset at each. We chose checkpoints obtained on steps
$\{99,199,\ldots,999\}$ and split each interval between two adjacent checkpoints into 10 pieces.

\end{document}